\documentclass[10pt,journal,compsoc]{IEEEtran}
\ifCLASSOPTIONcompsoc
  \usepackage[nocompress]{cite}
\else
  \usepackage{cite}
\fi
\ifCLASSINFOpdf
\else
\fi
\usepackage{algorithm}
\usepackage{algorithmic}

\usepackage{url}

\usepackage{graphicx}
\usepackage{amsmath}
\usepackage{amssymb}
\usepackage{booktabs}

\usepackage[hidelinks]{hyperref}
\usepackage{multirow}
\usepackage{subcaption}
\usepackage{bbm}
\usepackage{enumitem}
\usepackage{paralist}

\usepackage[capitalize]{cleveref}
\crefname{section}{Sec.}{Secs.}
\Crefname{section}{Section}{Sections}
\Crefname{table}{Table}{Tables}
\crefname{table}{Tab.}{Tabs.}

\newcommand{\etal}{\textit{et al}. }
\newcommand{\eg}{\textit{e.g}.}
\newcommand{\ie}{\textit{i.e}.}

\usepackage{pgfplots}
\usepackage{tikz}
\usepackage{tikzscale}
\usepackage[utf8]{inputenc}
\DeclareUnicodeCharacter{2212}{−}
\usepgfplotslibrary{groupplots,dateplot}
\usetikzlibrary{patterns,shapes.arrows}
\pgfplotsset{compat=newest}

\begin{document}
%
\title{Scalable SoftGroup for 3D Instance Segmentation on Point Clouds}
%
%
%
%

\author{Thang~Vu,~\IEEEmembership{Member,~IEEE,}
	Kookhoi~Kim,~\IEEEmembership{Member,~IEEE,}
	Thanh~Nguyen,~\IEEEmembership{Member,~IEEE,}
	Tung~M.~Luu,~\IEEEmembership{Member,~IEEE,}
	Junyeong~Kim,~\IEEEmembership{Member,~IEEE,}
    and~Chang~D.~Yoo,~\IEEEmembership{Senior~Member,~IEEE}
\IEEEcompsocitemizethanks{\IEEEcompsocthanksitem T. Vu, K. Kim, T. Nguyen, T. M. Luu, and C.D. Yoo are with the School of Electrical Engineering, Korea Advanced Institute of Science and Technology, Republic of Korea, 34141. E-mail: \{thangvubk, rlarnrghlapz, tungluu2203, thanhnguyen, cd\_yoo\}@kaist.ac.kr

\IEEEcompsocthanksitem J. Kim is with Department of AI, Chung-Ang University, Republic of Korea, 06974. E-mail: junyeongkim@cau.ac.kr
\IEEEcompsocthanksitem Corresponding author: C. D. Yoo
}
}

%
%

\markboth{IEEE TRANSACTIONS ON PATTERN ANALYSIS AND MACHINE INTELLIGENCE (TPAMI)}%
{Shell \MakeLowercase{\textit{et al.}}: Bare Demo of IEEEtran.cls for Computer Society Journals}
%



\IEEEtitleabstractindextext{%
\begin{abstract}
	This paper considers a network referred to as SoftGroup for accurate and scalable 3D instance segmentation. Existing state-of-the-art methods produce hard semantic predictions followed by grouping instance segmentation results. Unfortunately, errors stemming from hard decisions propagate into the grouping, resulting in poor overlap between predicted instances and ground truth and substantial false positives.
 To address the abovementioned problems, SoftGroup allows each point to be associated with multiple classes to mitigate the uncertainty stemming from semantic prediction. It also suppresses false positive instances by learning to categorize them as background. Regarding scalability, the existing fast methods require computational time on the order of tens of seconds on large-scale scenes, which is unsatisfactory and far from applicable for real-time. Our finding is that the $k$-Nearest Neighbor ($k$-NN) module, which serves as the prerequisite of grouping, introduces a computational bottleneck. SoftGroup is extended to resolve this computational bottleneck, referred to as SoftGroup++. The proposed SoftGroup++ reduces time complexity with octree $k$-NN and reduces search space with class-aware pyramid scaling and late devoxelization. Experimental results on various indoor and outdoor datasets demonstrate the efficacy and generality of the proposed SoftGroup and SoftGroup++. Their performances surpass the best-performing baseline by a large margin (6\% $\sim$ 16\%) in terms of AP$_{50}$. On datasets with large-scale scenes, SoftGroup++ achieves a 6$\times$ speed boost on average compared to SoftGroup. Furthermore, SoftGroup can be extended to perform object detection and panoptic segmentation with nontrivial improvements over existing methods. 
\end{abstract}

}

\maketitle

\IEEEdisplaynontitleabstractindextext

%
\IEEEpeerreviewmaketitle

\IEEEraisesectionheading{\section{Introduction}\label{sec:intro}}

\IEEEPARstart{W}{ith} the rapid evolution of 3D sensors and the widespread availability of large-scale 3D datasets, there has been a notable surge in interest towards achieving a deeper understanding of 3D scenes. Instance segmentation on point clouds is a 3D perception task that serves as the foundation for a wide range of applications such as autonomous driving, virtual reality, and robot navigation. Instance segmentation processes the point clouds to output a category and an instance mask for each detected object. 

\begin{figure}[t]
	\includegraphics[width=\columnwidth]{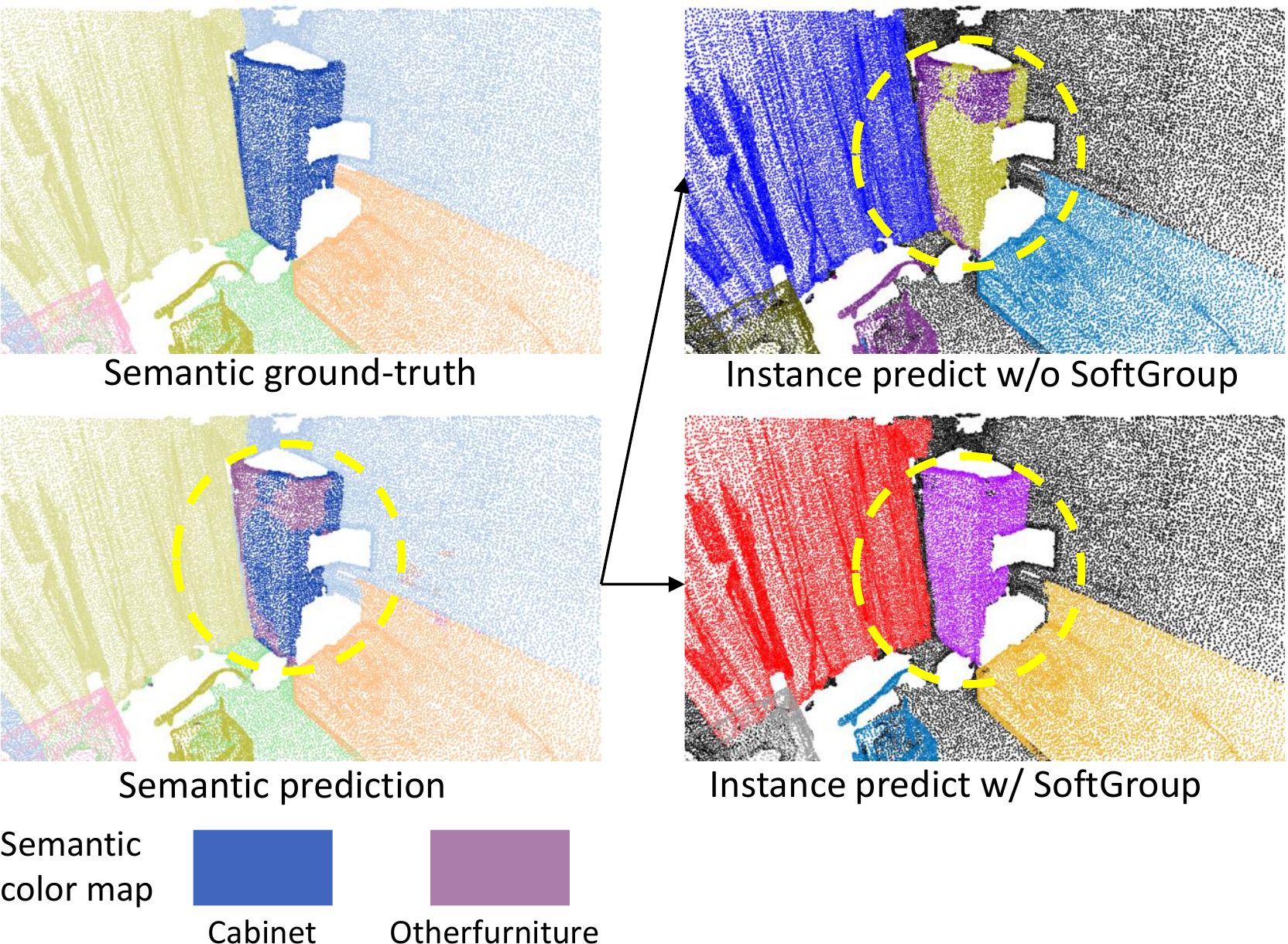}
	\caption{\textbf{Instance segmentation with and without SoftGroup from the same semantic prediction results.} The last row shows the palette for semantic predictions only. Instance predictions are illustrated by different random colors for different objects. In the semantic prediction results, some regions of \texttt{cabinet} are wrongly predicted as \texttt{other furniture}. Without SoftGroup, these errors are propagated to instance prediction. SoftGroup addresses this problem and produces more accurate instance masks.}
	\label{fig:introduction}
\end{figure}

This paper proposes a network referred to as SoftGroup for accurate and scalable point cloud instance segmentation. To attain high accuracy, existing state-of-the-art methods \cite{jiang2020pointgroup,liang2021instance,chen2021hierarchical} consider 3D instance segmentation as a bottom-up pipeline. These methods first predict the point-wise semantic labels and center offset vectors and then group points of the same predicted labels with small geometric distances into instances. These grouping algorithms are performed based on hard semantic predictions, where a point is associated with a single class. 
Unfortunately, objects frequently exhibit local ambiguity, and semantic segmentation tends to be noisy, often resulting in varied predictions for different parts of the same object. Consequently, employing hard semantic predictions for the purpose of instance grouping gives rise to two challenges: (1) the low overlap between the predicted instances and the ground truth, and (2) the introduction of additional false-positive instances stemming from incorrect semantic delineations.
\Cref{fig:introduction} shows a visualization example. Here, in the semantic prediction results, some parts of \texttt{cabinet} are wrongly predicted as \texttt{other furniture}. When hard semantic predictions are used to perform grouping, the semantic prediction error is propagated to instance prediction. As a result, the predicted \texttt{cabinet} instance has low overlap with the ground truth, and the \texttt{other furniture} instance is a false positive. 
The proposed SoftGroup overcomes these problems by considering soft semantic scores to perform grouping instead of hard one-hot semantic predictions. The intuition of SoftGroup is illustrated in \Cref{fig:intuation}. Our observation is that the object parts with incorrect semantic predictions still have reasonable scores for the true semantic class. SoftGroup relies on a score threshold instead of the maximum argument value in determining which category the object belongs to. Grouping on the soft semantic scores produces a more accurate instance of the true semantic class. The instance with incorrect semantic prediction will be suppressed by learning to categorize it as background. To this end, we treat an instance proposal (\ie, output of grouping) as either a positive or negative sample depending on its maximum Intersection over Union (IoU) with the ground truth. This is followed by a top-down refinement stage to refine the positive samples and suppress the negative samples. As shown in \Cref{fig:introduction}, SoftGroup can produce accurate instance masks from imperfect semantic prediction. 

SoftGroup is further extended for scalability to maintain fast inference speed on large-scale scenes. This advanced architecture will henceforth be referred to as SoftGroup++. Figure \ref{fig:runtime_vs_num_points} shows that the runtimes of existing methods grow quickly as the number of points increases. For instance, existing methods require  $\sim$20s to $\sim$50s to process a scene with $\sim$4.5M points. The processing time of each network component for this scene is further analyzed, as provided in Figure \ref{fig:component_time}. The results reveal that $k$-NN is the computational bottleneck, leading to the quick growth of inference time with regard to input size in HAIS \cite{chen2021hierarchical} and SoftGroup. These methods perform vanilla $k$-NN that requires measuring pair-wise distances of all points leading to time complexity of $\mathcal{O}(n^2)$ which is not scalable.

To enable fast inference on large-scale scenes, SoftGroup++ introduces an inference algorithm with low time complexity and search space. SoftGroup++ performs octree $k$-NN instead of vanilla $k$-NN to reduce time complexity from $\mathcal{O}(n^2)$ to $\mathcal{O}(n\log n)$. To effectively parallelize octree $k$-NN on GPU, a strategy to unfold the recursive structure of octree is presented such that tree traversal can be performed based on a simple arithmetic procedure. To reduce search space, SoftGroup++ relies on class-aware pyramid scaling and late devoxelization, which respectively perform adaptive downsampling on the backbone output features and delay the conversion from voxels to points until the end of the model. Figure \ref{fig:runtime_vs_num_points} shows that SoftGroup++ still maintains a fast inference speed as input size increases. The computational bottleneck of $k$-NN is also addressed, as presented in Figure \ref{fig:component_time}. 

\begin{figure}
	\centering
	\includegraphics[width=\columnwidth]{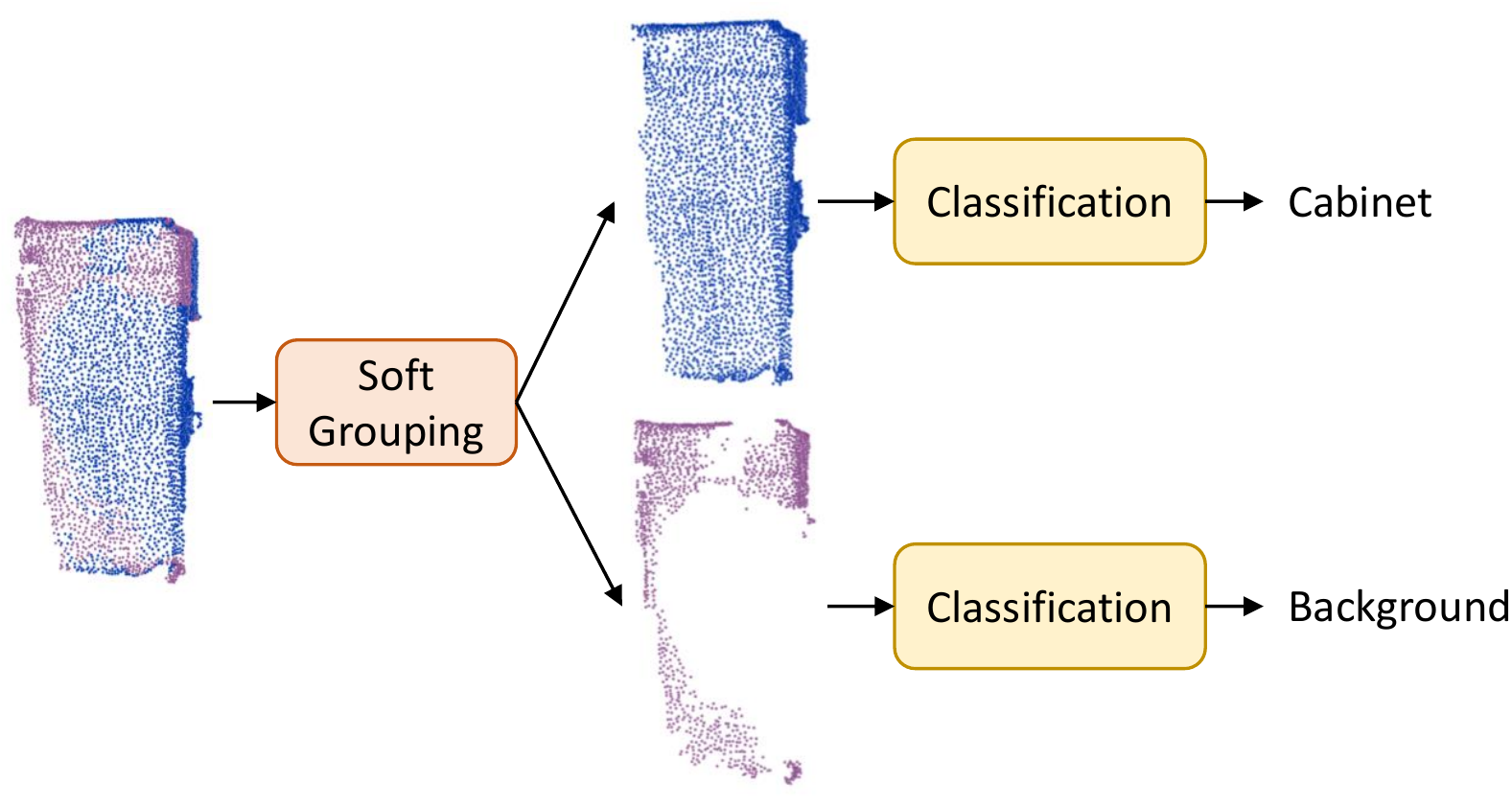}
	\caption{The \texttt{cabinet} in \Cref{fig:introduction} is extracted to illustrate the high-level pipeline of our method. The soft grouping module is based on soft semantic scores to output a more accurate instance (the upper one). The classifier processes each instance and suppresses the instance from wrong semantic prediction (the lower one).}
	\label{fig:intuation}
\end{figure}
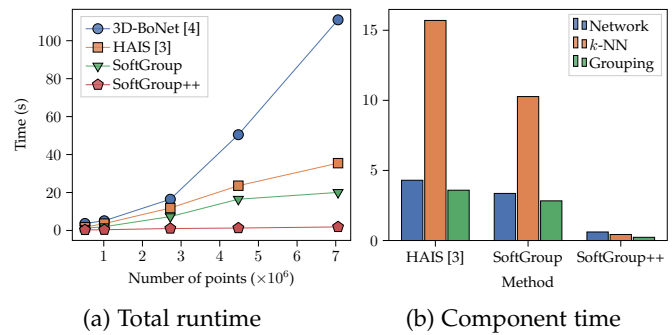
\begin{figure}
    \begin{subfigure}{0.485\columnwidth}
    \centering
    \begin{tikzpicture}[scale=0.54]
    \tikzstyle{every node}=[font=\fontsize{10.5pt}{10.5pt}\selectfont]
    \definecolor{color0}{rgb}{0.298039215686275,0.447058823529412,0.690196078431373}
    \definecolor{color1}{rgb}{0.866666666666667,0.517647058823529,0.32156862745098}
    \definecolor{color2}{rgb}{0.333333333333333,0.658823529411765,0.407843137254902}
    \definecolor{color3}{rgb}{0.768627450980392,0.305882352941176,0.32156862745098}
    
    \begin{axis}[
    legend cell align={left},
    legend style={
    	fill opacity=0.8,
    	draw opacity=1,
    	text opacity=1,
    	at={(0.03,0.97)},
    	anchor=north west,
    	draw=white!80!black
    },
    tick align=outside,
    tick pos=left,
    x grid style={white!69.0196078431373!black},
    xlabel={Number of points ($\times10^6$)},
    xmin=0.1863577, xmax=7.3873163,
    xtick style={color=black},
    y grid style={white!69.0196078431373!black},
    ylabel={Time (s)},
    ymin=-5.25075, ymax=116.53575,
    ytick style={color=black}
    ]
    \addplot [semithick, color0, mark=*, mark size=3.5, mark options={solid,draw=black}]
    table {%
    	0.513674 3.69
    	1.012866 5.1
    	2.721241 16.45
    	4.481021 50.46
    	7.06 111
    };
    \addlegendentry{3D-BoNet \cite{yang2019learning}}
    \addplot [semithick, color1, mark=square*, mark size=3.3, mark options={solid,draw=black}]
    table {%
    	0.513674 1.75
    	1.012866 3.61
    	2.721241 11.83
    	4.481021 23.59
    	7.06 35.47
    };
    \addlegendentry{HAIS \cite{chen2021hierarchical}}
    \addplot [semithick, color2, mark=triangle*, mark size=4, mark options={solid,rotate=180,draw=black}]
    table {%
    	0.513674 1.14
    	1.012866 1.97
    	2.721241 7.3
    	4.481021 16.46
    	7.06 20.05
    };
    \addlegendentry{SoftGroup}
    \addplot [semithick, color3, mark=pentagon*, mark size=4, mark options={solid,draw=black}]
    table {%
    	0.513674 0.285
    	1.012866 0.386
    	2.721241 1.02
    	4.481021 1.272
    	7.06 1.866
    };
    \addlegendentry{SoftGroup++}
    \end{axis}
    
    \end{tikzpicture}
    \caption{Total runtime}
    \label{fig:runtime_vs_num_points}
    \end{subfigure}
    \begin{subfigure}{0.52\columnwidth}
    \centering
    \begin{tikzpicture}[scale=0.54]
    \tikzstyle{every node}=[font=\fontsize{10.5pt}{10.5pt}\selectfont]
    \definecolor{color0}{rgb}{0.298039215686275,0.447058823529412,0.690196078431373}
    \definecolor{color1}{rgb}{0.866666666666667,0.517647058823529,0.32156862745098}
    \definecolor{color2}{rgb}{0.333333333333333,0.658823529411765,0.407843137254902}
    
    \begin{axis}[
    legend cell align={left},
    legend style={fill opacity=0.8, draw opacity=1, text opacity=1, draw=white!80!black},
    tick align=outside,
    tick pos=left,
    x grid style={white!69.0196078431373!black},
    xlabel={Method},
    xmin=-0.5015, xmax=2.5015,
    xtick style={color=black},
    xtick={0,1,2},
    xticklabels={HAIS \cite{chen2021hierarchical},SoftGroup,SoftGroup++},
    y grid style={white!69.0196078431373!black},
    ymin=0, ymax=16.485,
    ytick style={color=black}
    ]
    \draw[draw=black,fill=color0] (axis cs:-0.365,0) rectangle (axis cs:-0.135,4.3);
    \addlegendimage{ybar,ybar legend,draw=black,fill=color0};
    \addlegendentry{Network}
    
    \draw[draw=black,fill=color0] (axis cs:0.635,0) rectangle (axis cs:0.865,3.36);
    \draw[draw=black,fill=color0] (axis cs:1.635,0) rectangle (axis cs:1.865,0.61);
    \draw[draw=black,fill=color1] (axis cs:-0.115,0) rectangle (axis cs:0.115,15.7);
    \addlegendimage{ybar,ybar legend,draw=black,fill=color1};
    \addlegendentry{$k$-NN}
    
    \draw[draw=black,fill=color1] (axis cs:0.885,0) rectangle (axis cs:1.115,10.27);
    \draw[draw=black,fill=color1] (axis cs:1.885,0) rectangle (axis cs:2.115,0.43);
    \draw[draw=black,fill=color2] (axis cs:0.135,0) rectangle (axis cs:0.365,3.59);
    \addlegendimage{ybar,ybar legend,draw=black,fill=color2};
    \addlegendentry{Grouping}
    
    \draw[draw=black,fill=color2] (axis cs:1.135,0) rectangle (axis cs:1.365,2.83);
    \draw[draw=black,fill=color2] (axis cs:2.135,0) rectangle (axis cs:2.365,0.23);
    \end{axis}
    
    \end{tikzpicture}
    
    \caption{Component time}
    \label{fig:component_time}
    \end{subfigure}
    \caption[Caption]{(a) \textbf{Total runtime.} The runtimes of existing methods significantly increase as the number of points increases. (b) \textbf{Component time.} We measure the component time of processing a large scene of $\sim$4.5M points. The measurement exposes $k$-NN as a computational bottleneck. }
\end{figure}

Both SoftGroup and SoftGroup++ are conceptually simple and easy to implement. For instance segmentation task, they outperform the previous state-of-the-art method by a large margin, ranging from 6\% to 16\% in terms of AP$_{50}$ on different indoor and outdoor datasets. SoftGroup++ shows scalability advantages on S3DIS with large-scale scenes with 6$\times$ inference speed up. The versatility of SoftGroup is also demonstrated by the extension to object detection and panoptic segmentation  with nontrivial improvements over existing methods.

\vspace{0.5em}\noindent \textbf{Differences from our conference paper.} This manuscript is a significant extension of our conference version, which was previously published in \cite{vu2022softgroup}. In this updated version, we further investigate the scalability of recent 3D instance segmentation methods and reveal the computational bottleneck of $k$-NN. To efficiently process large-scale scenes, we propose SoftGroup++ which aims to reduce time complexity and search space. Octree $k$-NN replaces vanilla $k$-NN to reduce the time complexity from $\mathcal{O}(n^2)$ to $\mathcal{O}(n\log n)$. Class-aware pyramid scaling and late devoxelization reduce the search space and runtime of intermediate network components. Our proposed method is extensively benchmarked on various tasks and datasets demonstrating its versatility and generality. The source code and trained models are available at \url{https://github.com/thangvubk/SoftGroup}.

%

\section{Related work}
\textbf{Deep Learning on 3D Point Clouds.} Point cloud representation is a common data format for 3D scene understanding. To process point clouds, early methods \cite{aubry2011wave,rusu2009fast,rusu2008aligning,bronstein2010scale} extract hand-crafted features based on statistical properties of points. Recent deep learning methods learn to extract features from points. PointNet-based methods \cite{qi2017pointnet,qi2017pointnet++} propose to process points through shared Multi-Layer Perceptron (MLP) and then aggregate regional and global features from symmetric functions, such as max-pooling. Convolution methods are actively explored for point cloud processing. Continuous convolution methods \cite{xu2018spidercnn,liu2019relation,wu2019pointconv,thomas2019kpconv} learn the kernels which are associated with the spatial distribution of local points. Discrete convolution methods \cite{hua2018pointwise,li2018pointcnn,graham20183d,choy20194d,maturana2015voxnet,riegler2017octnet} learn the kernels which are regular grids obtaining from point quantization. Transformers \cite{lee2019set,zhao2021point,park2022fast} and graph-based methods \cite{simonovsky2017dynamic,shen2018mining,wang2019dynamic} are also proposed to address the data irregularity of point clouds.

\vspace{0.5em}\noindent \textbf{Proposal-based Instance Segmentation.}
Proposal-based methods consider a top-down strategy that generates region proposals and then segments the object within each proposal. Existing proposal-based methods for 3D point clouds are highly influenced by the success of Mask-R CNN \cite{he2017mask} for 2D images. To handle data irregularity of point clouds,  Li \etal \cite{yi2019gspn} propose GSPN, which takes an analysis-by-synthesis strategy to generate high-objectness 3D proposals, which are refined by a region-based PointNet. Hou \etal \cite{hou20193d} present 3DSIS that combines multi-view RGB input with 3D geometry to predict bounding boxes and instance masks. Yang \etal \cite{yang2019learning} propose 3D-BoNet which directly outputs a set of bounding boxes without anchor generation and non-maximum suppression, then segments the object by a pointwise binary classifier. Liu \etal \cite{liu2020learning} present GICN to approximate the instance center of each object as a Gaussian distribution, which is sampled to get object candidates and then produce bounding boxes and instance masks.

\begin{figure*}
	\centering
	\includegraphics[width=\textwidth]{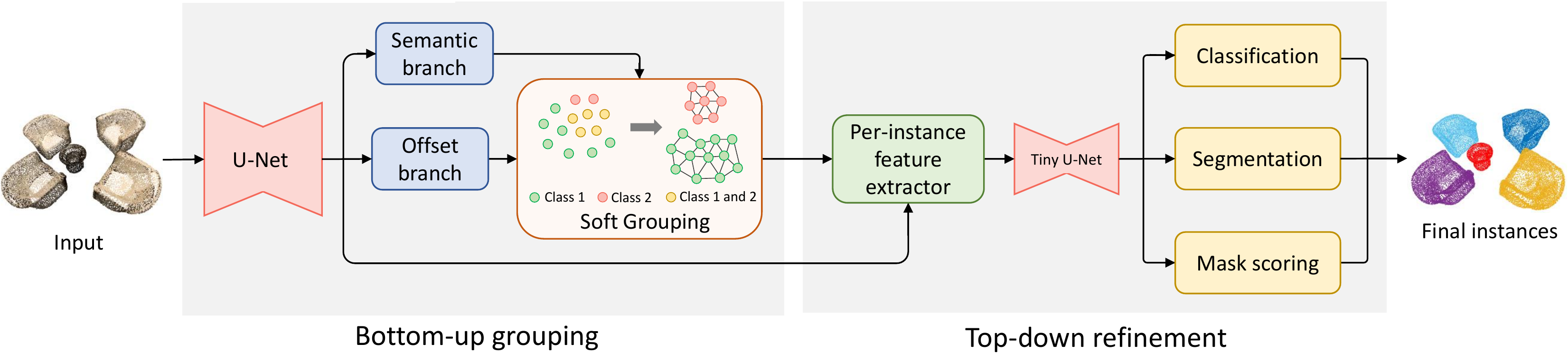}
	\caption{\textbf{The architecture of the proposed method consists of bottom-up grouping and top-down refinement stages.} From the input point clouds, the U-Net backbone extracts the point features. Then semantic and offset branches predict the semantic scores and offset vectors, followed by a soft grouping module to generate instance proposal. The soft grouping module addresses the problem of ambiguous semantic predictions, denoted by yellow points by assigning them to possible classes based on semantic scores for instance proposal generation. The feature extractor layer extracts backbone features from instance proposals. The features for each proposal are fed into a tiny U-Net followed by the classification, segmentation, and mask scoring branches to get the final instances.}
	\label{fig:arch}
\end{figure*}

\vspace{0.5em}\noindent \textbf{Grouping-based Instance Segmentation.}
Grouping-based methods rely on a bottom-up pipeline that produces per-point predictions (such as semantic maps, and geometric shifts, or latent features) then groups points into instances. Wang \etal \cite{wang2018sgpn} propose SGPN to construct a feature similarity matrix for all points and then group points of similar features into instances. Pham \etal \cite{pham2019jsis3d} present JSIS3D that incorporates the semantic and instance labels by a multi-value conditional random field model and jointly optimizes the labels to obtain object instances. Lahoud \etal \cite{lahoud20193d} propose MTML to learn feature and directional embedding, then perform mean-shift clustering on the feature embedding to generate object segments, which are scored according to their direction feature consistency. Han \etal \cite{han2020occuseg} introduce OccuSeg which performs graph-based clustering guided by object occupancy signal for more accurate segmentation outputs. Zhang \etal \cite{zhang2021point} consider a probabilistic approach that represents each point as a tri-variate normal distribution followed by a clustering step to obtain object instances. Jiang \etal \cite{jiang2020pointgroup} propose PointGroup to segment objects on original and offset-shifted point sets, relying on a simple yet effective algorithm that groups nearby points of the same label and expands the group progressively. Chen \etal \cite{chen2021hierarchical} extend PointGroup and propose HAIS that further absorbs surrounding fragments of instances and then refines the instances based on intra-instance prediction. Liang \etal \cite{liang2021instance} SSTNet to construct a tree network from pre-computed superpoints then traverse the tree and split nodes to get object instances. 

\vspace{0.5em}\noindent \textbf{Panoptic Segmentation.} Since most of the instance segmentation methods rely on semantic segmentation to produce instance results, these methods can be naturally extended to panoptic segmentation. Zhou \etal \cite{zhou2021panoptic} propose Panoptic-PolarNet to perform semantic segmentation and class-agnostic instance clustering on polar Bird's Eye View (BEV) representation for efficient panoptic segmentation.  Gasperini \etal \cite{gasperini2021panoster} introduce Panoster which is a learning-based solution object clustering, enabling direct derivation of panoptic results from semantic and instance predictions without post-processing. Hong \etal \cite{hong2021lidar} present DS-Net with Dynamic Shift module for object clustering, followed by consensus-driven fusion module to incorporate semantic and instance results. Li \etal \cite{li2022panoptic} propose Panoptic-PHNet with different improvements including fused 2D-3D backbone, transformer-based offset branch, and center grouping module for more accurate panoptic segmentation.


\vspace{0.5em}\noindent \textbf{Efficient Point Cloud Processing.} Due to data irregularity, efficient point cloud processing is crucial for various tasks, ranging from recognition to compression, and reconstruction. Riegler \etal \cite{riegler2017octnet} propose an octree-based convolutional network that hierarchically partitions the space to focus on relevant regions with low memory allocation and computation. Fu \etal \cite{fu2022octattention} presents an octree-based attention network for point cloud compression that extends the receptive field of context and exploits features from sibling nodes and their ancestors to model the dependency of nodes in large-scale context. Xu \etal \cite{xu2020grid} introduce a Grid Graph Convolutional Network (Grid-GCN) that utilizes the advantages of volumetric models and point-based models to achieve efficient data structuring and efficient computation at the same time. Rosu \etal \cite{rosu2020latticenet} propose LatticeNet for 3D semantic segmentation that embeds point features into a sparse permutohedral lattice for fast convolutions while keeping a low memory footprint. Lombardi \etal \cite{lombardi2020scalable} also utilize a permutohedral lattice on hierarchical point cloud features for efficient point cloud reconstruction. Park \etal \cite{park2022fast} proposes Fast Point Transformer with a lightweight self-attention layer and hashing-based architecture. 

The proposed SoftGroup and SoftGroup++ harness the benefits of both proposal-based and grouping-based approaches synergistically. They are constructed as a two-stage pipeline, where the bottom-up stage generates high-quality object proposals by grouping on soft semantic scores, and then the top-down stage process each proposal to refine positive samples and suppress negative ones. To efficiently process large-scale scenes, SoftGroup++ proposes to reduce time complexity and search space with octree $k$-NN, class-aware pyramid scaling, and late devoxelization.

\section{SoftGroup for Accurate Point Cloud Instance Segmentation}
This subsection presents SoftGroup with a focus on accuracy. The overall architecture of SoftGroup is depicted in Figure \ref{fig:arch}, which is divided into two stages. In the bottom-up grouping stage, the point-wise prediction network takes point clouds as the input and produces point-wise semantic labels and offset vectors. The soft grouping module processes these outputs to produce preliminary instance proposals. In the top-down refinement stage, based on the proposals, the corresponding features from the backbone are extracted and used to predict classes, instance masks, and mask scores as the final results.

\subsection{Point-wise Prediction Network}
\label{ssec:point_wise_net}
The input of the point-wise prediction network is a set of $N$ points, each of which is represented by its coordinate and color. The point set is voxelized to convert points into sparse volumetric grids, which are used as the input of a U-Net style backbone \cite{ronneberger2015u} to obtain point features. The Submanifold Sparse Convolution \cite{graham20183d} is adopted to implement the U-Net for 3D point clouds. The employed backbone architecture is identical to that of HAIS \cite{chen2021hierarchical}. From the output backbone features, the inverse mapping of the input voxelization step is applied to obtain point features. Then, two branches are constructed to produce the point-wise semantic scores and offset vectors.

\vspace{0.5em}\noindent \textbf{Semantic Branch.} A semantic branch is constructed from a two-layer MLP, which learns to output semantic scores $\boldsymbol{S} = \{\boldsymbol{s}_1, ..., \boldsymbol{s}_N\} \in \mathbb{R}^{N\times N_\text{class}}$ for $N$ points over $N_\text{class}$ classes. Different from existing methods \cite{jiang2020pointgroup,chen2021hierarchical}, we directly perform grouping on semantic scores without converting them into one-hot semantic predictions. 

\vspace{0.5em}\noindent \textbf{Offset Branch.} In parallel with the semantic branch, we apply a two-layer MLP to learn the offset vectors $\boldsymbol{O} = \{\boldsymbol{o}_1, ..., \boldsymbol{o}_N\} \in \mathbb{R}^{N\times 3}$, which represents the vector from each point to the geometric center of the instance the point belongs. Based on the learned offset vectors, we shift the points to the center of the corresponding instance to perform grouping more effectively.

The cross-entropy loss and $\ell_1$ regression loss are used to train the semantic and offset branches, respectively.

\begin{equation}
L_{\text{semantic}} = \frac{1}{N} \sum_{i=1}^{N}\text{CE}(\boldsymbol{s}_i, {s}^*_i),
\end{equation}
\begin{equation}
L_{\text{offset}} = \frac{1}{\sum_{i=1}^{N} \mathbbm{1}_{\{\boldsymbol{p}_i\}}} \sum_{i=1}^{N} \mathbbm{1}_{\{\boldsymbol{p}_i\}} \Vert \boldsymbol{o}_i - \boldsymbol{o}_i^* \Vert_1,
\end{equation}
where ${s}^*$ is the semantic label, $\boldsymbol{o}^*$ is offset label representing the vector from a point to the geometric center of the instance that the point belongs to (analogous to \cite{jiang2020pointgroup,chen2021hierarchical,liang2021instance}), and $\mathbbm{1}_{\{\boldsymbol{p}_i\}}$ is the indicator function indicating whether the point $\boldsymbol{p}_i$ belongs to any instance.
\subsection{Soft Grouping}
\label{ssec:softgroup}
The soft grouping module receives the semantic scores and offset vectors as the input and produces instance proposals. First, the offset vectors are used to shift points toward the corresponding instance centers. To perform grouping using the semantic scores, we define a score threshold $\tau$ to determine which semantic classes a point belongs to, allowing the point to be associated with multiple classes. Given semantic scores $\boldsymbol{S} \in \mathbb{R}^{N\times N_\text{class}}$, we iterate through $N_\text{class}$ classes, and at each class index we slice a point subset of the whole scene that has the score (w.r.t. the class index) higher than the threshold $\tau$. We follow \cite{jiang2020pointgroup,chen2021hierarchical} to perform grouping on each point subset. Since all points in each subset belong to the same class, we simply traverse all the points in the subset and create the links between points having a geometric distance smaller than a grouping radius $r$ to get the instance proposals. For each iteration, the grouping is performed on a point subset of the whole scan, ensuring fast inference. The overall instance proposals are the union of the proposals from all subsets.

We note that existing proposal-based methods \cite{hou20193d,yang2019learning,liu2020learning} commonly consider bounding boxes as object proposals then perform segmentation within each proposal. Intuitively, the bounding box that largely overlaps the instance should have the center close to the object center. However, generating high-quality bounding box proposals in 3D point clouds is challenging since the point only exists on object surfaces. Instead, SoftGroup relies on point-level proposals which are more accurate and naturally inherit the scattered property of point clouds. 

\begin{figure}
	\centering
	\begin{subfigure}{0.495\columnwidth}
		\includegraphics[width=\textwidth]{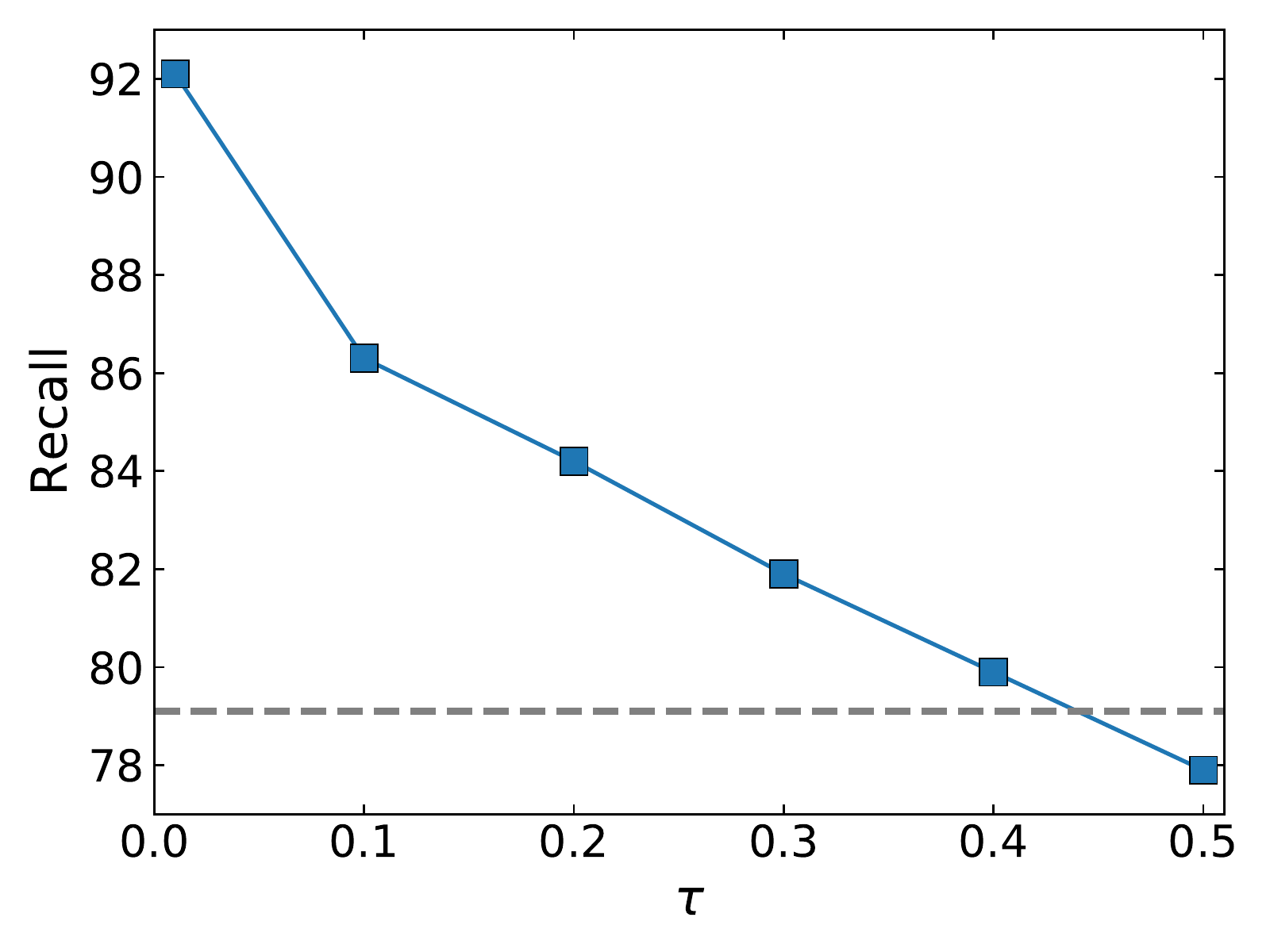}
	\end{subfigure}
	\begin{subfigure}{0.495\columnwidth}
		\includegraphics[width=\textwidth]{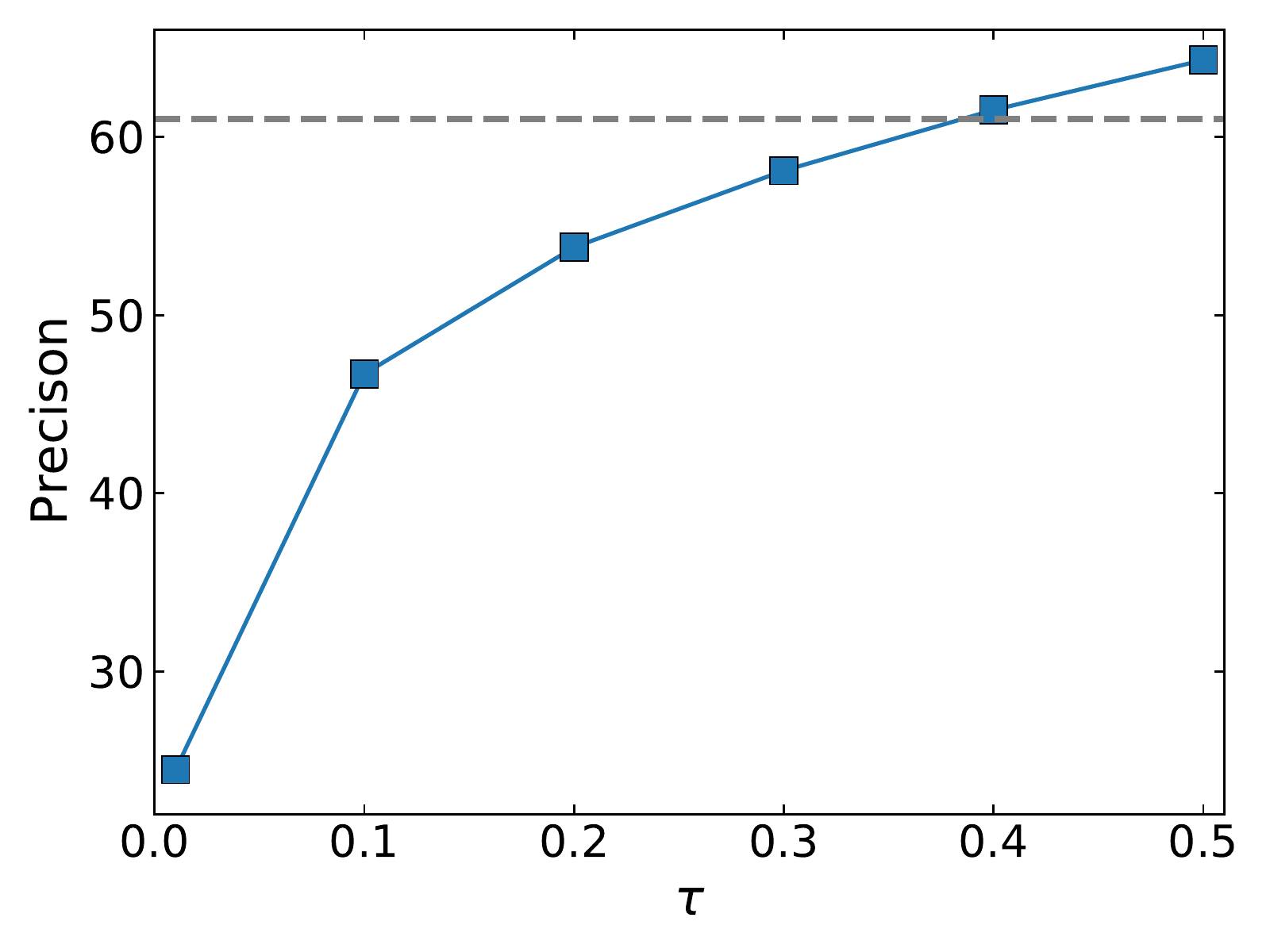}
	\end{subfigure}
	\caption{\textbf{The recall and precision of semantic prediction with varying score threshold $\tau$.} The dashed lines denote the recall and precision with hard semantic prediction.}
	\label{fig:recall_precision}
\end{figure}

Since the quality of instance proposals from grouping highly depends on the quality of semantic segmentation, we quantitatively analyze the impact of $\tau$ on the recall and precision of semantic predictions. The recall and precision for class $j$ are defined as follows.
\begin{equation}
\begin{aligned}
\text{recall}_j &= \sum_{i=1}^{N} \frac{(s_{ij} > \tau) \land (s^*_i = j)}{{s^*_i = j}},\\
\text{precision}_j &= \sum_{i=1}^{N} \frac{(s_{ij} > \tau) \land (s^*_i = j)}{s_{ij} > \tau}.
\end{aligned} 
\end{equation}

\Cref{fig:recall_precision} shows the recall and precision (averaged over classes) with the varying score thresholds $\tau$ compared with those of hard semantic prediction. With hard semantic prediction, the recall is 79.1\%, indicating more than 20\% amount of points over classes are not covered by the predictions. When using the score threshold, the recall increases as the score threshold decreases. However, the small score threshold also leads to low precision. We propose a top-down refinement stage to mitigate the low-precision problems. The precision can be interpreted as the relation between foreground and background points of object instances. We set the threshold to 0.2 with precision near 50\%, leading to the ratio between foreground and background points for the ensuring stage being balanced.

\subsection{Top-Down Refinement}
\label{ssec:top_down_refinement}
The top-down refinement stage classifies and refines the instance proposals from the bottom-up grouping stage. A feature extractor layer processes each proposal to extract its corresponding backbone features. The extracted features are fed into a tiny U-Net network (a U-Net style network with a small number of layers) before predicting classification scores, instance masks, and mask scores at the ensuing branches.

\vspace{0.5em}\noindent \textbf{Classification Branch.} The classification branch starts with a global average pooling layer to aggregate the feature of all points in the instance, followed by a MLP to predict the classification scores $\boldsymbol{C} = \{\boldsymbol{c}_1, ..., \boldsymbol{c}_K\} \in \mathbb{R}^{K\times (N_\text{class}+1)}$, where $K$ is the number of instances and $N_\text{class}+1$ indicates $N_{class}$ foreground classes with an extra background. We directly derive the object category and classification confidence score from the output of the classification branch. 

We note that existing grouping-based methods typically derive the object category from semantic predictions. However, instances may come from objects with noisy semantic predictions. The proposed method directly uses the output of the classification branch as the instance class. The classification branch aggregates all point features of the instance and classifies the instance with a single label, leading to more reliable predictions. 

\vspace{0.5em}\noindent \textbf{Segmentation Branch.}
As shown in \Cref{ssec:softgroup}, the instance proposals contain both foreground and background points, we construct a segmentation branch to predict an instance mask within each proposal. The segmentation branch is a point-wise MLP of two layers that output an instance mask $\boldsymbol{m}_k$  for each instance $k$.

\vspace{0.5em}\noindent \textbf{Mask Scoring Branch.}
The mask scoring branch shares the same structure as the classification branch. This branch outputs the mask scores $\boldsymbol{E} = \{\boldsymbol{e}_1, ..., \boldsymbol{e}_K\} \in \mathbb{R}^{K\times N_\text{class}}$, which estimate the IoU of a predicted mask with the ground truth. The mask score is combined with the classification score by multiplication to get the final confidence score. 

\begin{figure*}[t]
	\centering
	\begin{subfigure}{\textwidth}
		\centering
		\includegraphics[width=\textwidth]{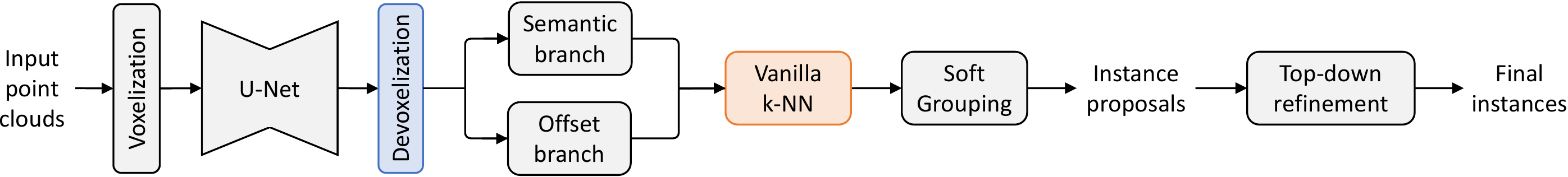}
		\caption{SoftGroup}
	\end{subfigure}
	
	\begin{subfigure}{\textwidth}
		\centering
		\includegraphics[width=\textwidth]{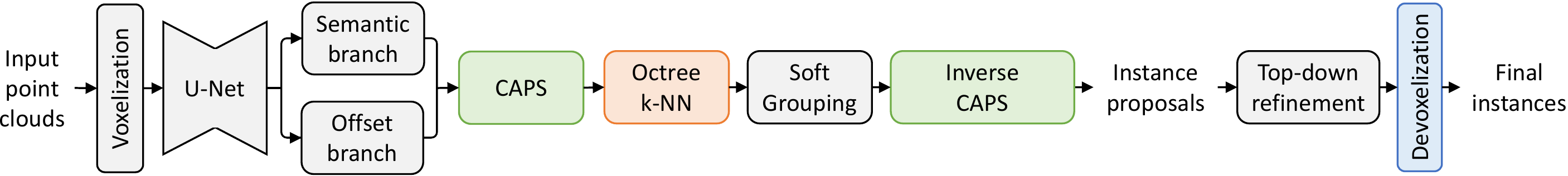}
		\caption{SoftGroup++}
	\end{subfigure}
	\caption{\textbf{Architecture comparison between SoftGroup and SoftGroup++.} To highlight the differences between them, devoxelization and vanilla k-NN of SoftGroup are shown. SoftGroup++ replaces vanilla $k$-NN with octree $k$-NN, introduces class-aware pyramid scaling (CAPS), and delay voxelization until the end of the network.}
     \label{fig:softgroup++}
\end{figure*}

\vspace{0.5em}\noindent \textbf{Learning Targets.} Training the top-down refinement branches requires the target labels for each branch. To this end, we follow the logic in existing 2D object detection and segmentation methods \cite{renNIPS15fasterrcnn,he2017mask}. We treat all instance proposals having IoU with a ground-truth instance higher than 50\% as the positive samples and the rest as negatives. Every positive sample is assigned to a ground-truth instance with the highest IoU. The classification target of a positive sample is the categorical label of the corresponding ground-truth instance. The segmentation and mask scoring branches are trained on positive samples only. The mask target of a positive sample is the mask of the assigned ground-truth instance. The mask score target is the IoU between the predicted mask and the ground truth. The training loss of these branches is the combination of cross-entropy, binary cross-entropy, and $\ell_2$ regression losses,  following \cite{he2017mask,huang2019mask}.

\begin{equation}
L_{\text{class}} = \frac{1}{K} \sum_{k=1}^{K}\text{CE}(\boldsymbol{c}_k, {c}^*_k),
\end{equation}
\begin{equation}
L_{\text{mask}} = \frac{1}{\sum_{k=1}^{K} \mathbbm{1}_{\{\boldsymbol{m}_k\}}} \sum_{k=1}^{K} \mathbbm{1}_{\{\boldsymbol{m}_k\}} \text{BCE}(\boldsymbol{m}_k, \boldsymbol{m}^*_k),
\end{equation}
\begin{equation}
L_{\text{mask\_score}} = \frac{1}{\sum_{k=1}^{K} \mathbbm{1}_{\{\boldsymbol{e}_k\}}} \sum_{k=1}^{K} \mathbbm{1}_{\{\boldsymbol{e}_k\}} \Vert \boldsymbol{e}_k - \boldsymbol{e}_k^* \Vert_2.
\end{equation}
Here, ${c}^*$, $\boldsymbol{m}^*$, $\boldsymbol{e}^*$ are the classification, segmentation, and mask scoring targets, respectively. $K$ is the total number of proposals and $\mathbbm{1}_{\{.\}}$ indicates whether the proposal is a positive sample. 

\subsection{Multi-task Learning}
The whole network can be trained in an end-to-end manner using a multi-task loss.
\begin{equation}
L = L_\text{semantic} + L_\text{offset} + L_\text{class} + L_\text{mask} + L_\text{mask\_score},
\end{equation}
where $L_\text{semantic}$ and $L_\text{offset}$ are the semantic and offset losses defined at \Cref{ssec:point_wise_net} while $L_\text{class}$, $L_\text{mask}$ and $L_\text{mask\_score}$ are the classification, segmentation and mask score losses defined at \Cref{ssec:top_down_refinement}.

\section{SoftGroup++ for Scalable Point Cloud Instance Segmentation}
For fast inference on large-scale scenes, SoftGroup++ extends SoftGroup with two major improvements: time complexity and search space reduction. The overall architecture of SoftGroup++ in comparison with SoftGroup is illustrated in Figure \ref{fig:softgroup++}. To highlight the differences between these two architectures, relevant components for comparison of SoftGroup (\ie, devoxelization and vanilla $k$-NN) are shown. SoftGroup++ replaces vanilla $k$-NN with octree $k$-NN to reduce time complexity from $\mathcal{O}(n^2)$ to $\mathcal{O}(n\log n)$. Additionally, class-aware pyramid scaling (CAPS) and late devoxelization are proposed to reduce search space. These improving components are parameter-free, and thus inference on SoftGroup and SoftGroup++ can be performed from the same trained models. Importantly, since each component can receive either point or voxel input, the presentation in the next subsections assumes point input is used for simplicity.


\subsection{Time Complexity Reduction}

\vspace{0.5em}\noindent \textbf{Octree $k$-NN.} In recent grouping-based instance segmentation methods \cite{jiang2020pointgroup,chen2021hierarchical}, $k$-NN constructs the point adjacency matrix, which serves as the prerequisite for grouping. To cope with varying point density, a radius constraint $r$ is added such that the distance from a valid neighbor to the query point should be less than $r$. Existing methods adopt vanilla $k$-NN algorithm, where pair-wise distance needs to be evaluated on the whole point set, and thus the time complexity of this algorithm w.r.t. the number of points is $\mathcal{O}(n^2)$. Since this quadratic time complexity is not scalable, we propose octree $k$-NN with time complexity of $\mathcal{O}(n\log n)$. 

\vspace{0.5em}\noindent \textbf{Constructing Octree.} Octree is a data structure that partitions the 3D space by recursively subdividing it into eight octants. Given a set of points, we first derive its tight axis-aligned bounding box. Then we recursively divide the 3D box into eight child boxes (octants). To balance the construction and traversal time, we limit the number of tree levels $M$ to a small value (\eg, 3) and store points in the last tree level (leaf nodes).

\begin{figure*}
	\centering
	\begin{subfigure}{0.4\textwidth}
		\centering
		\includegraphics[height=0.14\textheight]{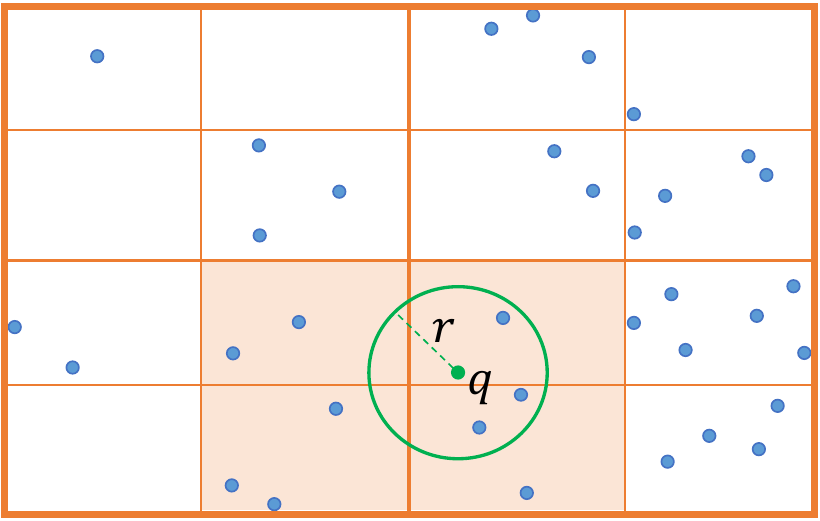}
		\caption{Quadtree $k$-NN}
		\label{fig:octree_search}
	\end{subfigure}
	\begin{subfigure}{0.56\textwidth}
		\centering
		\includegraphics[height=0.14\textheight]{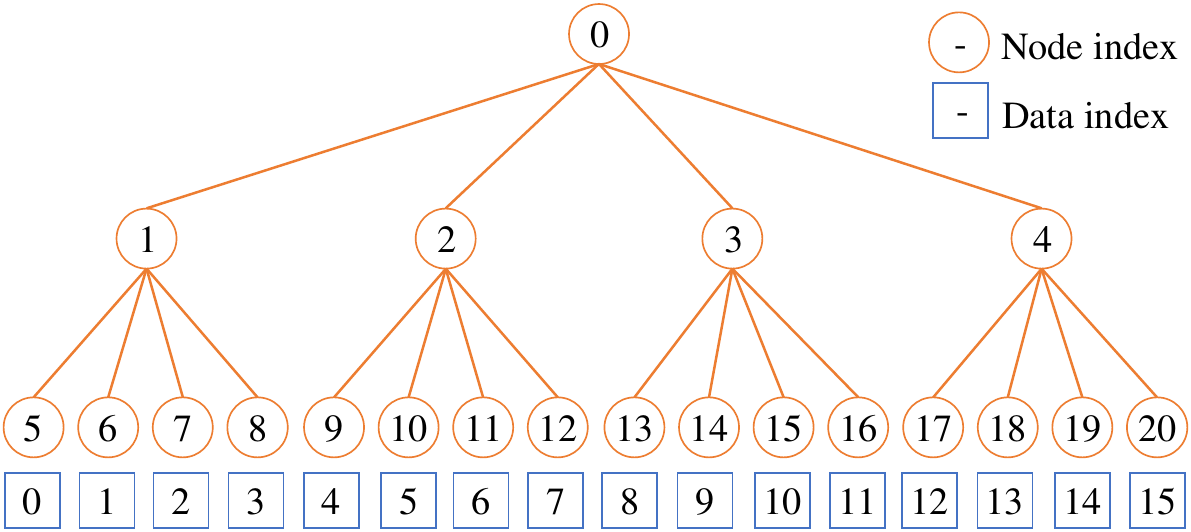}
		\caption{Node index and data index in breadth-first order}
		\label{fig:octree_graph}
	\end{subfigure}
	\caption{\textbf{Visualization of quadtree $k$-NN and breadth-first indices.} For neat presentation, we consider a quadtree of 2D data. (a) \textbf{Illustration of 2-level quadtree $k$-NN on query $q$ with search radius $r$.} There are 4 leaf nodes (highlighted in orange background) that intersects query circle. Quadtree $k$-NN only measures distance from query to the points in these leaf nodes instead of the whole point set. \textbf{(b) Illustration of node and data indices in breadth-first order.} From breadth-first order, the recursive structure of the tree can be unfolded where child nodes and data indices can be retrieved by simple arithmetic given in Eqs. \ref{eq:child} and \ref{eq:data} such that the tree-based search algorithm can be parallelized effectively on GPU.}
	\label{fig:octree}
\end{figure*}

\vspace{0.5em}\noindent \textbf{$k$-Nearest Neighbor Search on Octree.} Given query point $q$ and the constructed octree, we are ready to perform $k$-NN of the query $q$ with radius search $r$. The details are presented in Algorithm \ref{alg:algorithm}. The core idea of the algorithm is to find a small point subset near the query and then perform vanilla $k$-NN on the subset as opposed to the whole set. Starting from the root node, the algorithm recursively traverses through the tree. If the box associated with the current node intersects the sphere $S(q, r)$, there exist octants of the current node that intersects the sphere $S(q, r)$. These octants are enqueued and then checked in next iterations. The procedure is repeated until the leaf nodes. A point list $P$ is used to store all the points associated with the leaf nodes having intersections with the sphere $S(q, r)$. Figure \ref{fig:octree_search} illustrates the results of the algorithm on a two-level tree. For a neat presentation, we consider a quadtree which is the 2D version of the octree. The query sphere has intersections with 4 boxes of leaf nodes, of which the points are taken to perform $k$-NN with the query. 

\begin{algorithm}
	\begin{algorithmic}[1]
	\REQUIRE $q$: point query, $r$: search radius, $S(q, r)$: sphere with center $q$ and radius $r$, $\mathit{root}$: root node of constructed octree, each node is associated with corresponding boxes, points, and octants.
	\ENSURE $k$ nearest neighbors of query $q$ with radius $r$.
	\STATE Initialize an empty point list $P$
	\STATE Initialize an empty node queue $Q$
	\STATE $Q$.enqueue($\mathit{root}$)
	\WHILE{$Q$ is not empty}
	\STATE $\mathit{node}$ = $Q$.dequeue()
	\IF{$\mathit{node.box}$ $\cap$ $S(q,r)$ $\neq \varnothing$}
	\IF{$\mathit{node}$ is not leaf}
	\FOR{$octant$ in $node.octants$}
	\STATE $Q$.enqueue($\mathit{octant})$
	\ENDFOR
	\ELSE
	\STATE $P$.append($node.points$)
	\ENDIF
	\ENDIF
	\ENDWHILE
	\STATE Perform $k$-NN of query $q$ with radius $r$ on point set $P$
	\end{algorithmic}
	\caption{Octree $k$-NN}
	\label{alg:algorithm}
\end{algorithm}

\begin{algorithm*}
	\caption{Class-aware pyramid scaling}
	\label{alg:pyramid_scaling}
	\begin{algorithmic}[1]
	\REQUIRE $L$: number of pyramid levels, $V$: base voxel size, $C$: number of classes, $\boldsymbol{t} = [t_1, ..., t_{L}]$: increasing order thresholds to determine pyramid levels, $\boldsymbol{S}$: semantic predictions, $\boldsymbol{O}$: offset predictions.
	\ENSURE $\boldsymbol{S}'$: semantic predictions after CAPS, $\boldsymbol{O}'$: offset predictions after CAPS.
	\STATE Initialize empty sets $\boldsymbol{S}' = \{\}$ and $\boldsymbol{O}' = \{\}$\\
	\FOR{$i = 1$ \textbf{to} $C$}
	\STATE Extract semantic and offset prediction subsets for $i$-th class $\boldsymbol{S}_i \subset \boldsymbol{S}$ and $\boldsymbol{O}_i \subset \boldsymbol{O}$
	\STATE Get number of voxels in the subset $|\boldsymbol{S_i}|$\\
	\STATE Compute pyramid level: $l=\text{argmin}_{j\in \{1,\dots,L\}}\{t_j > |\boldsymbol{S_i}|\} $\\
	\STATE Voxelize $\boldsymbol{S}_i$ and $\boldsymbol{O}_i$ with voxel size being $l \times V$ which results in $\boldsymbol{S}'_i$  and $\boldsymbol{O}'_i$\\
	\STATE Aggregate downscaled results: $\boldsymbol{S}' =\boldsymbol{S}' \cup  \boldsymbol{S}'_i$ , $\boldsymbol{O}' =\boldsymbol{O}' \cup  \boldsymbol{O}'_i$
	\ENDFOR
	\end{algorithmic}
\end{algorithm*}

To construct the adjacency matrix of all points, the algorithm is performed on each point of the point set, which requires GPU parallelization for a speed boost. One of the main challenges for parallelization is that the implementation of recursive tree traversal on CUDA kernel is nontrivial. Inspired by \cite{miller2011real,riegler2017octnet}, we present a simple strategy to unfold and traverse the tree using direct indexing, such that the algorithm is performed on each CUDA kernel for each point effectively. To this end, when constructing the tree, we index the tree nodes in breadth-first order, as shown in Figure \ref{fig:octree_graph}, given a parent node index $i$, we can access its child nodes directly by simple arithmetic:
\begin{equation}
\text{ch}_j(i) = i \times 2^d + j \text{ for } j = 1..2^d
\label{eq:child}
\end{equation}
where $\text{ch}_j(i)$ is index of the $j$-th child node and $d$ is dimension ($d$ is 3 in octree or 2 in quadtree). Differ from \cite{miller2011real,riegler2017octnet}, we construct a full tree such that it is not required to encode the tree structure. The data index to access the points associated with the leaf node $i$ can be derived as:
\begin{equation}
\text{data}(i) = i - \sum_{m=0}^{M-1}2^{md}
\label{eq:data}
\end{equation}
Since node and data indices are derived via simple arithmetic, we can unfold the recursive structure and parallelize the octree $k$-NN such that each CUDA kernel performs the algorithm for a query.

\subsection{Search Space Reduction}

\vspace{0.5em}\noindent \textbf{Class-Aware Pyramid Scaling.} Class-Aware Pyramid Scaling (CAPS) is proposed to reduce search space for $k$-NN and soft grouping module. Here, the search space is referred to as the number of input points used in $k$-NN. In point cloud processing, common search space reduction methods are point sampling (\eg, random sampling, furthest point sampling) and voxelization. In this work, voxelization is adopted since point-level results can be simply attained using the inverse mapping of voxelization. It is important to note that the voxelization step of CAPS is independent of input voxelization at the beginning of the network. Before delving into the proposed CAPS, we present a naive scaling strategy for search space reduction. In particular, the naive scaling strategy directly voxelizes semantic and offset predictions of the whole scene with a single voxel size. The scores/features of points within a voxel are averaged to get corresponding voxel scores/features. This strategy exhibits two major limitations: (1) the scores of a class may interfere with the scores of other classes due to score averaging in voxelization and (2) it is difficult to choose a fixed voxel size for downscaling the whole scene since different object classes may have different object sizes. As ablated in our experiments (Section \ref{sec:softgroup++_ablation_study}), this strategy severely worsens the prediction accuracy.

The limitations of the naive scaling strategy are addressed in the proposed CAPS. To avoid score interference between classes, the proposed CAPS is class-aware such that downscaling is performed on a point subset for a given class. To cope with the variation in object sizes of different classes, CAPS adaptively selects downscaling levels such that the object with more points has a higher downscaling level. The details of CAPS are presented in Algorithm \ref{alg:pyramid_scaling}. The algorithm iterates through $C$ classes and extract semantic $\boldsymbol{S}_i$ and offset $\boldsymbol{O}_i$ subsets for $i$-th class. The pyramid level $l$ is computed by comparing the number of points in $\boldsymbol{S}_i$ with predefined thresholds $\boldsymbol{t}$. Then, $\boldsymbol{S}_i$ and $\boldsymbol{O}_i$ are downscaled with a voxel size of $l \times V$, where V is the base voxel size. The downscaled semantic and offset predictions of all classes are aggregated and used as the input for $k$-NN and soft grouping. Since top-down refinement stage requires instance proposals of the same scale, a module referred to as inverse CAPS is applied, where inverse mapping of the voxelization in CAPS is used to perform inverse scaling. 

    
    
    


\vspace{0.5em}\noindent \textbf{Late Devoxelization.} In voxel-based networks, it is common that the number of voxels is much less than the number of points. Table \ref{tab:voxel_vs_point} illustrates the ratio between voxels and points in different datasets. The number of voxels is computed with the voxel size according to Table \ref{tab:hyperparameter}. Existing methods \cite{jiang2020pointgroup,liang2021instance,chen2021hierarchical} typically perform early devoxelization, where the conversion from voxels to points is performed right after the U-Net backbone. It is noted that all the points within a voxel share the same values after devoxelization.  Early devoxelization lead to the repetition of computation on the same input, and thus the ensuing modules are operated at a high computational cost. To address this problem, this work proposes to delay the conversion from voxels to points until the end of the network, which is referred to as late devoxelization. Late devoxelization eliminates computation repetition and reduces input size for intermediate network components, leading to a speed boost. The effectiveness of late devoxelization can be anticipated by the voxel-to-point ratio. 

\begin{table}[t]
\centering
\caption{Ratio between \#voxels and \#points.}
\begin{tabular}{cccc}
\toprule
ScanNet v2 & S3DIS & STLS3D & SemanticKITTI\\ \midrule
0.69 & 0.27      &  0.92 & 0.72         \\ \bottomrule
\end{tabular}
\label{tab:voxel_vs_point}
\end{table}

\section{Experiments}
\subsection{Experimental Settings}
\vspace{0.5em}\noindent \textbf{Datasets.}
 The experiments are conducted on various indoor and outdoor datasets, including S3DIS \cite{armeni20163d}, ScanNet v2 \cite{dai2017scannet}, STPLS3D \cite{chen2022stpls3d}, SemanticKITTI \cite{behley2019semantickitti}. The ScanNet dataset contains 1613 scans which are divided into training, validation, and testing sets of 1201, 312, 100 scans, respectively. Instance segmentation is evaluated on 18 object classes. Following existing methods, the benchmarked results are reported on the hidden test split. The ablation study is conducted on the validation set.

The S3DIS dataset contains 3D scans of 6 areas with 271 scenes in total. The dataset consists of 13 classes for instance segmentation evaluation. Following existing methods, two settings are used to evaluate the instance segmentation results: testing on Area 5 and 6-fold cross-validation.

The STPLS3D dataset is recently proposed for outdoor 3D instance segmentation with more than 16km$^2$ of landscapes. The dataset is annotated with 14 instance segmentation classes. The point clouds are cropped into non-overlapped blocks of 250m$^2$, resulting in nearly 1 million points each scene. We follow \cite{chen2022stpls3d} to divide the data to train and test splits.

The SemanticKITTI dataset is proposed for panoptic segmentation, which unifies instance segmentation (for thing classes) and semantic segmentation (for stuff classes). The dataset consists of 22 sequences with 43552 frames. The training and validation are performed on sequences 00-10, and testing is conducted on sequences 11-21. Panoptic segmentation is evaluated on 11 thing and 8 stuff classes.

\vspace{0.5em}\noindent \textbf{Evaluation Metrics.} The evaluation metric is the standard average precision. Here, AP$_{50}$ and AP$_{25}$ denote the scores with IoU thresholds of 50\% and 25\%, respectively. Likewise, AP denotes the averaged scores with IoU threshold from 50\% to 95\% with a step size of 5\%. Additionally, the S3DIS is also evaluated using mean precision (mPrec), and mean recall (mRec), which are adopted in previous methods. For panpotic segmentation, the panoptic quality (PQ) \cite{kirillov2019panoptic} is used as the default evaluation metric. PQ captures both recognition quality (RQ) and segmentation quality (PQ). PQ$^{th}$ and PQ$^{st}$ indicate the scores on thing and stuff classes separately. In addition, PQ$^\dagger$ is also reported as suggested in \cite{porzi2019seamless}. Runtime is measured on a single RTX 8000. 

\begin{table}[]
\centering
\setlength{\tabcolsep}{3.2pt}
\caption{Voxel size and grouping bandwidth (in meters) for different datasets.}
\begin{tabular}{@{}lcccc@{}}
\toprule
Hyper-parameter    & ScanNet v2 & S3DIS & STLS3D & SemanticKITTI \\ \midrule
Voxel size         & 0.02       & 0.02  & 0.33   & 0.05          \\
Grouping bandwidth & 0.04       & 0.04  & 0.90   & 0.10          \\ \bottomrule
\end{tabular}
\label{tab:hyperparameter}
\end{table}

 \begin{table*}[]
	\small
	\centering
	\setlength{\tabcolsep}{2.8pt}
    \caption{3D instance segmentation results on ScanNet v2 hidden test set in terms of AP$_{50}$ scores.}
	\begin{tabular}{@{}l|c|cccccccccccccccccc@{}} \toprule
		Method         & AP$_{50}$ & \rotatebox[origin=c]{90}{bathtub} & \rotatebox[origin=c]{90}{bed } & \rotatebox[origin=c]{90}{bookshe.} & \rotatebox[origin=c]{90}{cabinet} & \rotatebox[origin=c]{90}{chair} & \rotatebox[origin=c]{90}{counter} & \rotatebox[origin=c]{90}{curtain} & \rotatebox[origin=c]{90}{desk} & \rotatebox[origin=c]{90}{door} & \rotatebox[origin=c]{90}{other} & \rotatebox[origin=c]{90}{picture} & \rotatebox[origin=c]{90}{fridge} & \rotatebox[origin=c]{90}{s. curtain} & \rotatebox[origin=c]{90}{sink} & \rotatebox[origin=c]{90}{sofa} & \rotatebox[origin=c]{90}{table} & \rotatebox[origin=c]{90}{toilet} & \rotatebox[origin=c]{90}{window} \\ \midrule
		SGPN \cite{wang2018sgpn}          & 14.3          & 20.8           & 39.0          & 16.9          & 6.5           & 27.5          & 2.9           & 6.9         & 0.0           & 8.7           & 4.3           & 1.4           & 2.7           & 0.0            & 11.2          & 35.1          & 16.8          & 43.8           & 13.8          \\
		GSPN \cite{yi2019gspn}          & 30.6          & 50.0           & 40.5          & 31.1          & 34.8          & 58.9          & 5.4           & 6.8         & 12.6          & 28.3          & 29.0          & 2.8           & 21.9          & 21.4           & 33.1          & 39.6          & 27.5          & 82.1           & 24.5          \\
		3D-SIS \cite{hou20193d}        & 38.2          & \textbf{100.0} & 43.2          & 24.5          & 19.0          & 57.7          & 1.3           & 26.3        & 3.3           & 32.0          & 24.0          & 7.5           & 42.2          & 85.7           & 11.7          & 69.9          & 27.1          & 88.3           & 23.5          \\
		MASC \cite{liu2019masc}          & 44.7          & 52.8           & 55.5          & 38.1          & 38.2          & 63.3          & 0.2           & 50.9        & 26.0          & 36.1          & 43.2          & 32.7          & 45.1          & 57.1           & 36.7          & 63.9          & 38.6          & 98.0           & 27.6          \\
		PanopticFusion \cite{narita2019panopticfusion} & 47.8          & 66.7           & 71.2          & 59.5          & 25.9          & 55.0          & 0.0           & 61.3        & 17.5          & 25.0          & 43.4          & 43.7          & 41.1          & 85.7           & 48.5          & 59.1          & 26.7          & 94.4           & 35.9          \\
		3D-Bonet \cite{yang2019learning}      & 48.8          & \textbf{100.0} & 67.2          & 59.0          & 30.1          & 48.4          & 9.8           & 62.0        & 30.6          & 34.1          & 25.9          & 12.5          & 43.4          & 79.6           & 40.2          & 49.9          & 51.3          & 90.9           & 43.9          \\
		MTML \cite{lahoud20193d}          & 54.9          & \textbf{100.0} & 80.7          & 58.8          & 32.7          & 64.7          & 0.4           & 81.5        & 18.0          & 41.8          & 36.4          & 18.2          & 44.5          & \textbf{100.0} & 44.2          & 68.8          & 57.1          & \textbf{100.0} & 39.6          \\
		3D-MPA \cite{engelmann20203d}        & 61.1          & \textbf{100.0} & 83.3          & 76.5          & 52.6          & 75.6          & 13.6          & 58.8        & 47.0          & 43.8          & 43.2          & 35.8          & 65.0          & 85.7           & 42.9          & 76.5          & 55.7          & \textbf{100.0} & 43.0          \\
		Dyco3D \cite{he2021dyco3d}        & 64.1          & \textbf{100.0} & 84.1          & 89.3          & 53.1          & 80.2          & 11.5          & 58.8        & 44.8          & 43.8          & 53.7          & 43.0          & 55.0          & 85.7           & 53.4          & 76.4          & 65.7          & 98.7           & 56.8          \\
		PE \cite{zhang2019point}            & 64.5          & \textbf{100.0} & 77.3          & 79.8          & 53.8          & 78.6          & 8.8           & 79.9        & 35.0          & 43.5          & 54.7          & 54.5          & 64.6          & 93.3           & 56.2          & 76.1          & 55.6          & 99.7           & 50.1          \\
		PointGroup \cite{jiang2020pointgroup}    & 63.6          & \textbf{100.0} & 76.5          & 62.4          & 50.5          & 79.7          & 11.6          & 69.6        & 38.4          & 44.1          & 55.9          & 47.6          & 59.6          & \textbf{100.0} & 66.6          & 75.6          & 55.6          & 99.7           & 51.3          \\
		GICN \cite{liu2020learning}          & 63.8          & \textbf{100.0} & \textbf{89.5} & 80.0          & 48.0          & 67.6          & 14.4          & 73.7        & 35.4          & 44.7          & 40.0          & 36.5          & 70.0          & \textbf{100.0} & 56.9          & 83.6          & 59.9          & \textbf{100.0} & 47.3          \\
		OccuSeg \cite{han2020occuseg}       & 67.2          & \textbf{100.0} & 75.8          & 68.2          & 57.6          & 84.2          & \textbf{47.7} & 50.4        & 52.4          & 56.7          & 58.5          & 45.1          & 55.7          & \textbf{100.0} & 75.1          & 79.7          & 56.3          & \textbf{100.0} & 46.7          \\
		SSTNet \cite{liang2021instance}        & 69.8          & \textbf{100.0} & 69.7          & 88.8          & 55.6          & 80.3          & 38.7          & 62.6        & 41.7          & 55.6          & 58.5          & \textbf{70.2} & 60.0          & \textbf{100.0} & \textbf{82.4} & 72.0          & 69.2          & \textbf{100.0} & 50.9          \\
		HAIS \cite{chen2021hierarchical}          & 69.9          & \textbf{100.0} & 84.9          & 82.0          & 67.5          & 80.8          & 27.9          & 75.7        & 46.5          & 51.7          & 59.6          & 55.9          & 60.0          & \textbf{100.0} & 65.4          & 76.7          & 67.6          & 99.4           & 56.0          \\
		\textbf{SoftGroup (ours)}    & 76.1          & \textbf{100.0} & 80.8          & 84.5          & \textbf{71.6} & 86.2          & 24.3          & 82.4        & 65.5          & \textbf{62.0} & 73.4          & 69.9          & 79.1          & 98.1           & 71.6          & 84.4          & 76.9          & \textbf{100.0} & 59.4 \\
		\textbf{SoftGroup++ (ours)}    & \textbf{76.9} & \textbf{100.0}   & 80.3          & \textbf{93.7} & 68.4          & \textbf{86.5} & 21.3          & \textbf{87.0} & \textbf{66.4} & 57.1          & \textbf{75.8} & \textbf{70.2} & \textbf{80.7} & \textbf{100.0}   & 65.3          & \textbf{90.2} & \textbf{79.2} & \textbf{100.0}   & \textbf{62.6} \\ \bottomrule
	\end{tabular}
	\label{tab:scannet_benchmark}
\end{table*}

\begin{table}[]
	\centering
	\small
    \caption{3D instance segmentation results on S3DIS dataset. Methods marked with ${\dagger}$ are evaluated on Area 5, and methods marked with  ${\ddagger}$ are evaluated on 6-fold cross-validation.}
	\begin{tabular}{l|ccccc}
		\toprule
		Method & AP            & AP$_{50}$          & mPrec$_{50}$       & mRec$_{50}$     \\ \midrule
		SGPN$^{\dagger}$ \cite{wang2018sgpn}       & -             & -             & 36            & 28.7            \\
		ASIS$^{\dagger}$ \cite{wang2019associatively}        & -             & -             & 55.3          & 42.4            \\
		PointGroup$^{\dagger}$ \cite{jiang2020pointgroup}  & -             & 57.8          & 61.9          & 62.1            \\
		SSTNet$^{\dagger}$ \cite{liang2021instance}      & 42.7          & 59.3          & 65.5          & 64.2           \\
		HAIS$^{\dagger}$ \cite{chen2021hierarchical}        & -             & -             & 71.1          & 65.0         \\
		\textbf{SoftGroup}$^{\dagger}$ \textbf{(ours)}   & \textbf{51.6} & 66.1          & 73.6          & 66.6         \\
		\textbf{SoftGroup++}$^{\dagger}$ \textbf{(ours)} & 50.9          & \textbf{67.8} & \textbf{73.8} & \textbf{67.6} \\ \midrule
		SGPN$^{\ddagger}$ \cite{wang2018sgpn}        & -             & -             & 38.2          & 31.2            \\
		ASIS$^{\ddagger}$ \cite{wang2019associatively}        & -             & -             & 63.6          & 47.5            \\
		3D-BoNet$^{\ddagger}$ \cite{yang2019learning}         & -             & -             & 65.6          & 47.7            \\
		PointGroup$^{\ddagger}$ \cite{jiang2020pointgroup}  & -             & 64.0            & 69.6          & 69.2           \\
		SSTNet$^{\ddagger}$ \cite{liang2021instance}      & 54.1          & 67.8          & 73.5          & 73.4          \\
		HAIS$^{\ddagger}$ \cite{chen2021hierarchical}        & -             & -             & 73.2          & 69.4        \\
		\textbf{SoftGroup}$^{\ddagger}$ \textbf{(ours)}   & 54.4          & 68.9          & 75.3          & 69.8          \\
		\textbf{SoftGroup++}$^{\ddagger}$ \textbf{(ours)} & \textbf{56.6} & \textbf{71.3} & \textbf{75.9} & \textbf{74.4} \\ \bottomrule
	\end{tabular}
	\label{tab:s3dis_benchmark}
\end{table}

\begin{table}[]
    \small
    \centering
    \caption{3D instance segmentation results on STPLS3D.}
    \begin{tabular}{l|ccc}
    \toprule
    Method   & AP   & AP$_{50}$ & AP$_{25}$ \\ \midrule 
    PointGroup \cite{jiang2020pointgroup}  & 23.3 & 38.5 & 48.6 \\
    HAIS \cite{chen2021hierarchical}       & 35.1 & 46.7 & 52.8 \\
    \textbf{SoftGroup (ours)}  & \textbf{47.3} & \textbf{63.1} & 71.4 \\
    \textbf{SoftGroup++ (ours)} & 46.5 & 62.9 & \textbf{71.8} \\ \bottomrule
    \end{tabular}
    \label{tab:stpls3d}
\end{table}

\begin{table}[]
	\small
	\centering
    \caption{Runtime comparison on a single RTX 8000. Numbers w/o and w/ parenthesizes are maximum and average runtime per scan, respectively.}
	\begin{tabular}{@{}l|ccc@{}} \toprule
		Method      & S3DIS                & ScanNet v2           & STPLS3D              \\ \midrule
        PointGroup \cite{jiang2020pointgroup} & -                    & 1.16 (0.30)          & -                    \\
        SSTNet \cite{liang2021instance}     & -                    & 1.56 (0.29)          & -                    \\
        HAIS \cite{chen2021hierarchical}       & 23.59 (3.87)         & 0.85 (0.27)          & 3.88 (2.83)          \\
        \textbf{SoftGroup (ours)}   & 16.46 (2.20)         & 0.84 (0.20)          & 3.42 (2.33)          \\
        \textbf{SoftGroup++ (ours)} & \textbf{1.27 (0.38)} & \textbf{0.41 (0.14)} & \textbf{2.24 (1.30)} \\ \bottomrule
	\end{tabular}

    \label{tab:runtime}
\end{table}

\vspace{0.5em}\noindent \textbf{Implementation Details.}
The implementation details follow those of existing methods \cite{jiang2020pointgroup,chen2021hierarchical}. The model is implemented using PyTorch deep learning framework \cite{paszke2017automatic} and trained by Adam optimizer \cite{kingma2014adam} with a batch size of 4. The learning rate is initialized to 0.001 and scheduled by a cosine annealing \cite{loshchilov2016sgdr}. The score threshold for soft grouping $\tau$ is set to 0.2. Note that the voxel size and grouping radius $r$ should be adapted for different datasets. We following \cite{jiang2020pointgroup, chen2022stpls3d} to set these hyper-parameters, as summarized in Table \ref{tab:hyperparameter}. Extra hyper-parameters of SoftGroup++ are set as follows. CAPS uses 3 levels with the threshold to determine pyramid level being $\boldsymbol{t} = [10^5, 10^6, +\infty]$. Based voxel size of CAPS is set to the input voxel size. The number of octree levels is set to 3. At training time, the scenes are randomly cropped to at most 250k points. At inference, the whole scene is fed into the network without cropping. 

Especially, S3DIS has large-scale scenes of high-point density, SoftGroup and SoftGroup++ are set up differently. In particular, SoftGroup follows \cite{jiang2020pointgroup} to randomly subsample the scenes at a ratio of 1/4. At inference, the scene is divided into four parts before feeding into the model, and then the features of these parts are merged right after the U-Net backbone. Meanwhile, since SoftGroup++ is efficient, the whole scene is used without subsampling. This implementation detail ensures consistency between training and testing for the network, thus bringing extra AP improvements of SoftGroup++ on S3DIS dataset.

\subsection{Benchmarking Results}
\subsubsection{Accuracy analysis}
\vspace{0.5em}\noindent \textbf{ScanNet v2.}
Table \ref{tab:scannet_benchmark} shows the results of our method and recent state-of-the-art on the hidden test set of ScanNet v2 benchmark. SoftGroup achieves AP$_{50}$ of 76.1\%, surpassing the existing best method (HAIS) by a significant margin of 6.2\%. SoftGroup++ slightly improves the performance to 76.9\%. Regarding class-wise scores, SoftGroup++ achieves the best performance in 13 out of 18 classes.

\begin{figure*}
	\centering
	
	\begin{subfigure}{0.195\textwidth}
		\includegraphics[width=\textwidth]{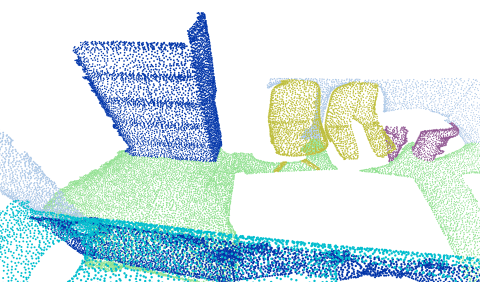}
	\end{subfigure}
	\begin{subfigure}{0.195\textwidth}
		\includegraphics[width=\textwidth]{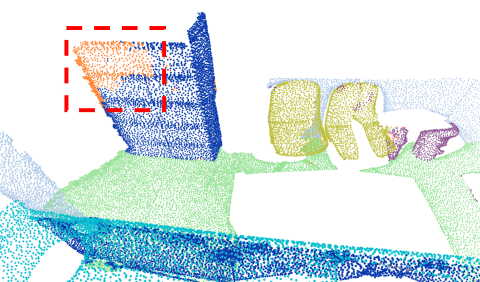}
	\end{subfigure}
	\begin{subfigure}{0.195\textwidth}
		\includegraphics[width=\textwidth]{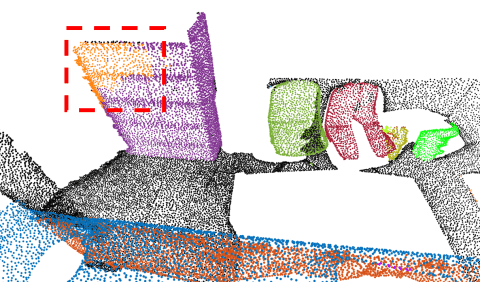}
	\end{subfigure}
	\begin{subfigure}{0.195\textwidth}
		\includegraphics[width=\textwidth]{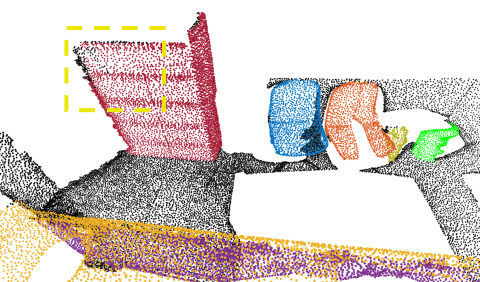}
	\end{subfigure}
	\begin{subfigure}{0.195\textwidth}
		\includegraphics[width=\textwidth]{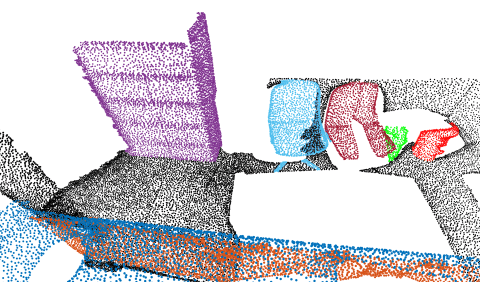}
	\end{subfigure}
	
	\begin{subfigure}{0.195\textwidth}
		\includegraphics[width=\textwidth]{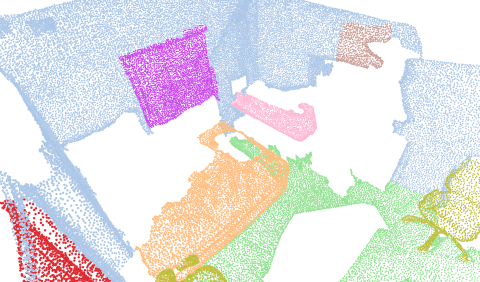}
		\caption*{Semantic GT}
	\end{subfigure}
	\begin{subfigure}{0.195\textwidth}
		\includegraphics[width=\textwidth]{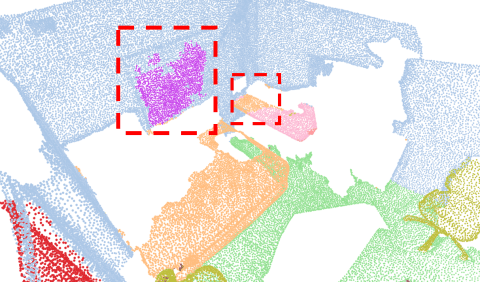}
		\caption*{Semantic pred}
	\end{subfigure}
	\begin{subfigure}{0.195\textwidth}
		\includegraphics[width=\textwidth]{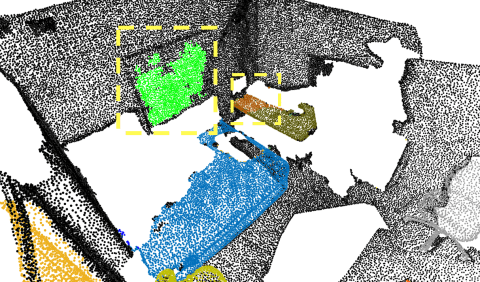}
		\caption*{Inst pred w/o SoftGroup}
	\end{subfigure}
	\begin{subfigure}{0.195\textwidth}
		\includegraphics[width=\textwidth]{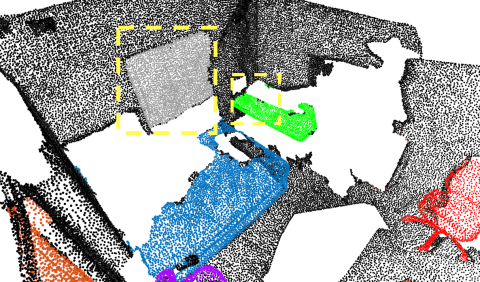}
		\caption*{Inst pred w/ SoftGroup}
	\end{subfigure}
	\begin{subfigure}{0.195\textwidth}
		\includegraphics[width=\textwidth]{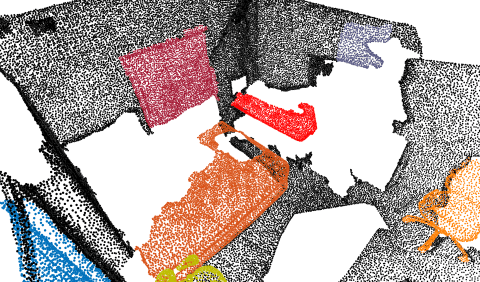}
		\caption*{Instance GT}
	\end{subfigure}
	\caption{\textbf{Qualitative results on ScanNet v2 validation set.} Instance prediction without SoftGroup output low-quality instance mask at the region of wrong semantic prediction (highlighted by dashed boxes). SoftGroup produces more accurate instance masks at these regions.}
	\label{fig:quanlitative_res}
\end{figure*}

\vspace{0.5em}\noindent \textbf{S3DIS.}
Table \ref{tab:s3dis_benchmark} summarizes the results on Area 5 and 6-fold cross-validation of S3DIS dataset. On both Area 5 and cross-validation evaluations, the proposed SoftGroup and SoftGroup++ achieve higher overall performance compared to existing methods. Notably, on Area 5 evaluation, SoftGroup achieves AP/AP$_{50}$ of 51.6/66.1(\%), which is 8.9/6.8(\%) improvement compared to SSTNet. SoftGroup++ further improves the overall results with 1.7\% and 2.4\% higher than SoftGroup on Area 5 and cross-validation settings, respectively.

\vspace{0.5em}\noindent \textbf{STPLS3D.} We report instance segmentation results on STPLS3D dataset in Table \ref{tab:stpls3d}. STPLS3D is an outdoor dataset with larger sparsity and object size variation compared to ScanNet and S3DIS. SoftGroup achieves AP/AP$_{50}$/AP$_{25}$ of 47.3/63.1/71.4(\%) which is 12.2/16.4/21.1(\%) improvement compared to HAIS. SoftGroup++ performs comparably to SoftGroup with  AP/AP$_{50}$/AP$_{25}$ of 46.5/62.9/71.8(\%). The significant performance on different indoor and outdoor datasets demonstrates the superiority and generalization of the proposed method.

\subsubsection{Runtime Analysis.}
Table \ref{tab:runtime} reports the runtime comparison on different dataset. For a fair comparison, the reported runtime is measured on the same RTX 8000 GPU model. Since the number of points for a scan varies, we report both maximum and average runtime. SoftGroup shows slight improvement compared to HAIS. SoftGroup++ achieves the fastest inference speed in all measures. Notably, on S3DIS which consists of large-scale scenes, SoftGroup++ requires a maximum runtime of 1.27s, which is significantly faster than SoftGroup of 16.46s. On average, SoftGroup++ is 6$\times$ faster than SoftGroup on S3DIS.

\subsection{SoftGroup Ablation Study}
\vspace{0.5em}\noindent \textbf{SoftGroup Component-wise Analysis.}
We provide experimental results of SoftGroup when different components are omitted. The considered baseline is a model with hard grouping and the confidence scores of output instances are ranked by a ScoreNet branch \cite{jiang2020pointgroup,liang2021instance}. Table \ref{tab:component_analysis} shows the ablation results. The baseline achieves 39.5/61.1/75.5(\%) in terms of AP/AP$_{50}$/AP$_{25}$. Significant improvement is obtained by either applying soft grouping or top-down refinement. Combining these two components achieves the best overall performance AP/AP$_{50}$/AP$_{25}$ of 46.0/67.6/78.9(\%), which is significantly higher than the baseline by 6.5/6.5/3.4(\%).

\begin{table}[!t]
	\small
	\centering
	\setlength{\tabcolsep}{2.0pt}
    \caption{Component-wise analysis on ScanNet v2 validation set. Our model achieves significant improvement over the baseline. }
	\begin{tabular}{@{}ccc|ccc@{}}
		\toprule
		Baseline & Soft grouping & Top-down refinement & AP   & AP$_{50}$ & AP$_{25}$ \\ \midrule
		\checkmark        &               &                 & 39.5 & 61.1 & 75.5 \\
		& \checkmark             &                 & 41.6 & 63.8 & \textbf{79.2} \\
		&               & \checkmark               & 44.3 & 65.4 & 78.1 \\
		& \checkmark             & \checkmark               & \textbf{46.0}   & \textbf{67.6} & 78.9 \\ \midrule
		\multicolumn{3}{l|}{\textbf{Overall improvement}} & \textbf{+6.5} & \textbf{+6.5} & \textbf{+3.4} \\ \bottomrule
	\end{tabular}
	\label{tab:component_analysis}
\end{table}
\begin{table}[!t]
	\centering
	\small
	\setlength{\tabcolsep}{12pt}
    \caption{Ablation experiments on varying score threshold $\tau$ for soft grouping. ``None" denotes the threshold is not used, and the hard semantic prediction is used for grouping.}
	\begin{tabular}{c|ccc}
		\toprule
		$\tau$  & AP            & AP$_{50}$ & AP$_{25}$        \\ \midrule
		None & 44.3          & 65.4          & 78.1          \\ \midrule
		0.01 & 40.1          & 58.5          & 69.2          \\
		0.1  & 45.3          & 66.5          & 78.5          \\
		0.2  & \textbf{46.0} & \textbf{67.6} & \textbf{78.9} \\
		0.3  & 45.2          & 66.8          & 78.5          \\
		0.4  & 44.7          & 46.1          & 78.3          \\
		0.5  & 43.9          & 64.8          & 77.7          \\ \bottomrule
	\end{tabular}
	\label{tab:score_thr}
\end{table}
\begin{table}[!t]
	\centering
    \small
    \caption{The impact of each branch in top-down refinement on ScanNet v2 validation set.}
	\begin{tabular}{ccc|ccc}
		\toprule
		Class & Mask& Mask score& AP          &AP$_{50}$ & AP$_{25}$          \\ \midrule
		\checkmark            &             &                   & 41.1        & 64.6          & \textbf{79.7} \\
		\checkmark            & \checkmark           &                   & 45.7        & \textbf{68.4} & 79.5          \\
		\checkmark            & \checkmark           & \checkmark                 & \textbf{46.0} & 67.6          & 78.9          \\ \bottomrule
	\end{tabular}
	\label{tab:top_down_refinement}
\end{table}

\begin{figure*}[t]
	\centering
	\begin{subfigure}{0.245\textwidth}
		\includegraphics[width=\textwidth]{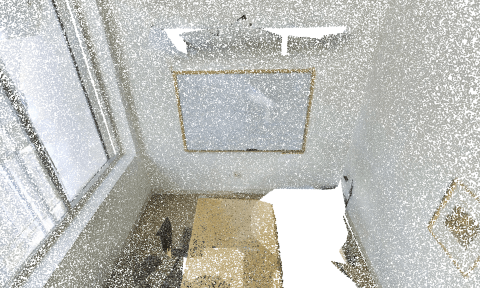}
	\end{subfigure}
	\begin{subfigure}{0.245\textwidth}
		\includegraphics[width=\textwidth]{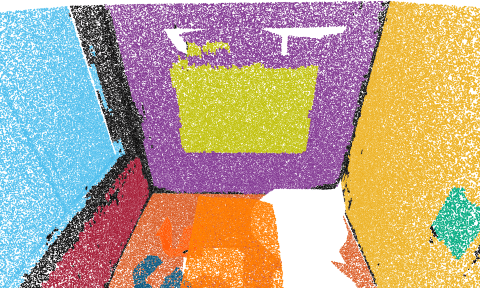}
	\end{subfigure}
	\begin{subfigure}{0.245\textwidth}
		\includegraphics[width=\textwidth]{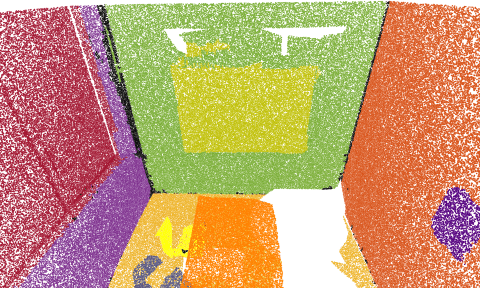}
	\end{subfigure}
	\begin{subfigure}{0.245\textwidth}
		\includegraphics[width=\textwidth]{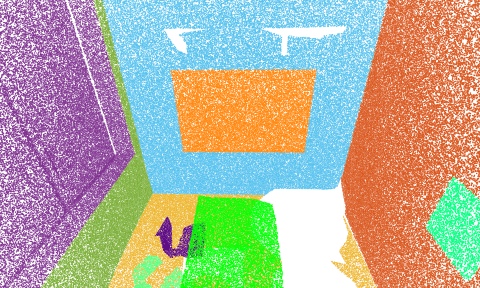}
	\end{subfigure}
    \vspace{0.05cm}
	
	\begin{subfigure}{0.245\textwidth}
		\includegraphics[width=\textwidth]{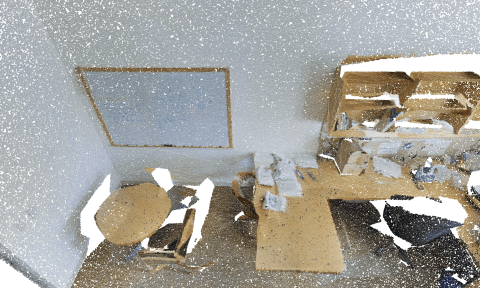}
		\caption{Input}
	\end{subfigure}
	\begin{subfigure}{0.245\textwidth}
		\includegraphics[width=\textwidth]{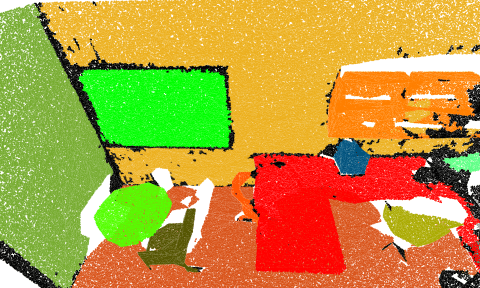}
		\caption{Naive scaling}
	\end{subfigure}
	\begin{subfigure}{0.245\textwidth}
		\includegraphics[width=\textwidth]{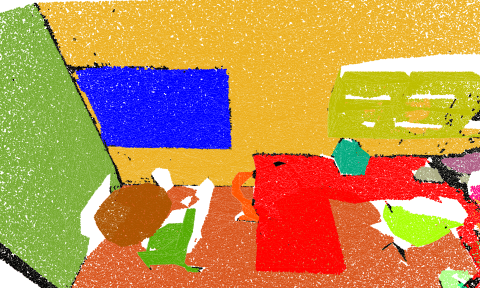}
		\caption{CAPS}
	\end{subfigure}
	\begin{subfigure}{0.245\textwidth}
		\includegraphics[width=\textwidth]{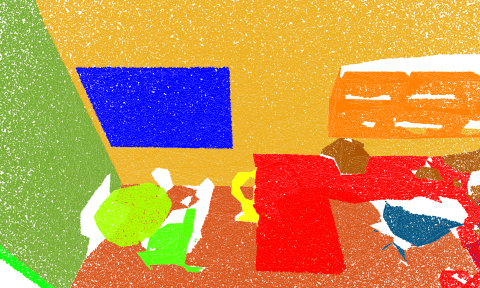}
		\caption{Ground truth}
	\end{subfigure}
	\caption{\textbf{Qualitative comparison between naive scaling and the proposed CAPS.} Naive scaling misdetects the instance masks around object edges, illustrated in black points. The proposed CAPS produces more accurate instance masks.}
	\label{fig:pyramid_scaling}
\end{figure*}

\begin{table*}[t]
	\centering
	\small
    \caption{Ablation study on each component of the proposed method on the largest scene of S3DIS Area 5 with \char`\~4.5M points. The module-wise and total runtimes (in seconds) are measured on RTX 8000.}
	\begin{tabular}{cccc|cccc|c}
		\toprule
		Baseline     &  Octree & CAPS &  Late devoxel. & Point-wise & $k$-NN  & Grouping & Top-down & Total \\ \midrule
		$\checkmark$ &                 &              &                     & 0.13       & 23.32 & 4.54     & 2.33                & 30.32 \\
  	$\checkmark$ & $\checkmark$    &              &                     & 0.13       & 2.82  & 4.54     & 2.36                & 9.85  \\
		$\checkmark$ &                 & $\checkmark$ &                     & 0.13       & 1.32  & 0.11     & 2.38                & 3.94  \\
		$\checkmark$ &                 &              & $\checkmark$        & 0.13       & 0.96  & 0.45     & 0.51                & 2.05  \\
		$\checkmark$ & $\checkmark$    & $\checkmark$ &                     & 0.13       & 1.28  & 0.11     & 2.35                & 3.87  \\
        $\checkmark$ &                 & $\checkmark$ & $\checkmark$        & 0.13	     & 0.50	&  0.23	    & 0.48	              & 1.34  \\
		$\checkmark$ & $\checkmark$    & $\checkmark$ & $\checkmark$        & 0.13       & 0.44  & 0.23     & 0.48                & \textbf{1.28}  \\ \bottomrule
	\end{tabular}

	\label{tab:component_wise}
\end{table*}

\vspace{0.5em}\noindent \textbf{Score Threshold for Soft Grouping.}
Table \ref{tab:score_thr} shows the experimental results with varying score thresholds for soft grouping. The baseline is with $\tau$ being ``None", indicating the threshold is deactivated and the hard predicted label is used for grouping. The baseline achieves AP/AP$_{50}$/AP$_{25}$ of 44.3/65.4/78.1(\%). When $\tau$ is too high or too low the performance is even worse than the baseline. The best performance is obtained at $\tau$ of 0.2, which confirms our analysis at the \Cref{ssec:softgroup}, where the number of positive and negative samples are balanced.

\vspace{0.5em}\noindent \textbf{Top-Down Refinement.}
We further provide the ablation results on the top-down refinement, on \Cref{tab:top_down_refinement}. With only the classification branch, our method achieves AP/AP$_{50}$/AP$_{25}$ of 41.1/64.6/79.7(\%). When mask branch and mask scoring branch are in turn applied, the performance tends to improve on the higher IoU threshold regions. Combining all branches yields the performance  AP/AP$_{50}$/AP$_{25}$ of 46.0/67.6/78.9(\%).



\vspace{0.5em}\noindent \textbf{Qualitative Analysis.} Figure \ref{fig:quanlitative_res} shows the visualization on ScanNet v2 dataset. Without SoftGroup, the semantic prediction errors are propagated to instance segmentation predictions (highlighted by dashed boxes). In contrast, SoftGroup effectively corrects the semantic prediction errors and thus generates more accurate instance masks.

\subsection{SoftGroup++ Ablation Study}
\label{sec:softgroup++_ablation_study}

\begin{table}[t]
\small
\centering
\caption{Ablation w/ and w/o octree k-NN on (uncropped) STPLS3D dataset.}
\setlength{\tabcolsep}{3pt}
\begin{tabular}{@{}c|cccc@{}}
\toprule
w/ Octree & Point-wise network & k-NN & Grouping & Top-down \\ \midrule
N         & 0.25               & 1.48 & 2.98     & 1.53     \\
Y         & 0.25               & 0.93 & 2.96     & 1.58     \\ \bottomrule
\end{tabular}
\label{tab:octree}
\end{table}

\begin{table}[t]
    \centering
    \small
    \caption{Ablation study of CAPS on S3DIS Area 5.}
	\begin{tabular}{c|cccc}
		\toprule
		Approach        & AP   & AP$_{50}$ & AP$_{25}$ & Time (s) \\ \midrule
		Naive scaling   & 47.0 & 64.7 & 73.7 & 0.44 \\
		CAPS & \textbf{50.9} & \textbf{67.8} & \textbf{76.9} & \textbf{0.39} \\ \bottomrule
	\end{tabular}
	\label{tab:pyramid_scaling}
\end{table}

\vspace{0.5em}\noindent \textbf{SoftGroup++ Component-wise Analysis.} The ablation study on SoftGroup++ is conducted on S3DIS since it consists of large-scale scenes. We report the experimental results when different components are omitted in Table \ref{tab:component_wise}. Without proposed components, the runtime of the baseline model is 30.32s with $k$-NN being the computational bottleneck of 23.32s. When octree $k$-NN, CAPS, and late devoxelization are individually applied, the runtime of the above bottleneck significantly to 2.82s, 1.32s, and 0.96s, respectively. When all proposed components are applied, the model achieves the lowest latency of 1.28s, which is nearly $24\times$ faster than the baseline.

\begin{figure*}
	\centering
	
	\begin{subfigure}{0.195\textwidth}
		\includegraphics[width=\textwidth]{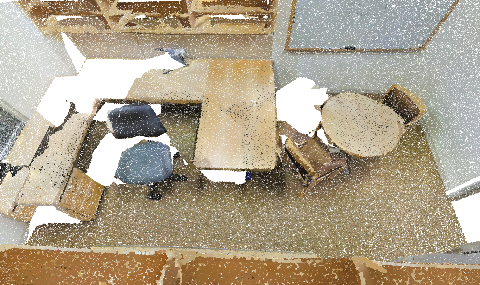}
            \caption*{Input}
	\end{subfigure}
	\begin{subfigure}{0.195\textwidth}
		\includegraphics[width=\textwidth]{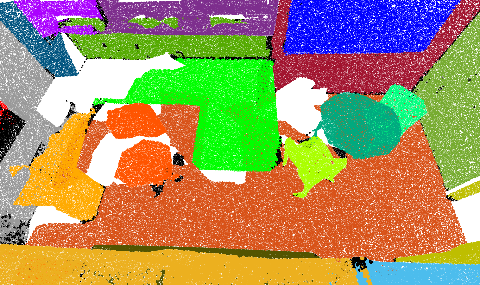}
            \caption*{Instance pred 1x}
	\end{subfigure}
	\begin{subfigure}{0.195\textwidth}
		\includegraphics[width=\textwidth]{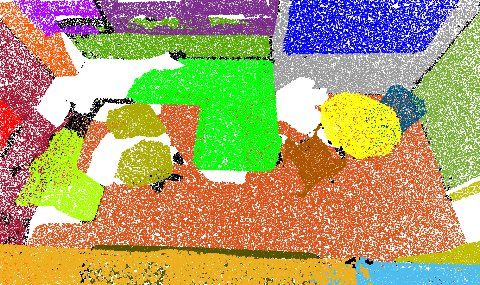}
            \caption*{Instance pred 0.5x}
	\end{subfigure}
	\begin{subfigure}{0.195\textwidth}
		\includegraphics[width=\textwidth]{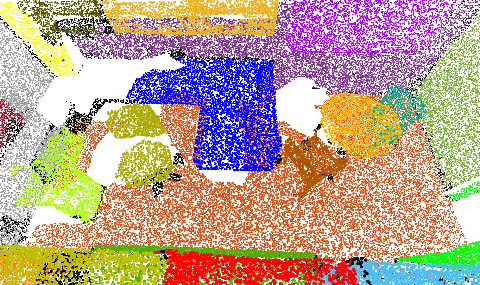}
            \caption*{Instance pred 0.25x}
	\end{subfigure}
	\begin{subfigure}{0.195\textwidth}
		\includegraphics[width=\textwidth]{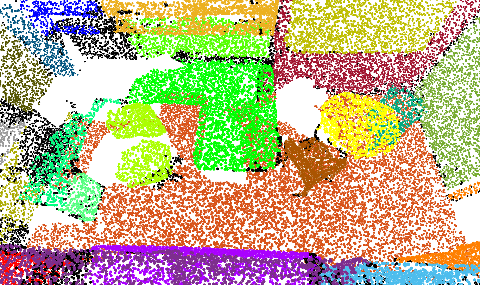}
            \caption*{Instance pred 0.125x}
	\end{subfigure}
	
	\caption{\textbf{Visualization examples of instance segmentation results with varying point density.} Instances are denoted by different random colors}
        \label{fig:point_density}
\end{figure*}

\begin{table*}[]
\small
\centering
\caption{Semantic segmentation results on ScanNet v2 validation set with back fusion.}
\setlength{\tabcolsep}{3.8pt}
\begin{tabular}{@{}l|c|cccccccccccccccccc@{}} \toprule
Method         & mIoU & \rotatebox[origin=c]{90}{bathtub} & \rotatebox[origin=c]{90}{bed } & \rotatebox[origin=c]{90}{bookshe.} & \rotatebox[origin=c]{90}{cabinet} & \rotatebox[origin=c]{90}{chair} & \rotatebox[origin=c]{90}{counter} & \rotatebox[origin=c]{90}{curtain} & \rotatebox[origin=c]{90}{desk} & \rotatebox[origin=c]{90}{door} & \rotatebox[origin=c]{90}{other} & \rotatebox[origin=c]{90}{picture} & \rotatebox[origin=c]{90}{fridge} & \rotatebox[origin=c]{90}{s. curtain} & \rotatebox[origin=c]{90}{sink} & \rotatebox[origin=c]{90}{sofa} & \rotatebox[origin=c]{90}{table} & \rotatebox[origin=c]{90}{toilet} & \rotatebox[origin=c]{90}{window} \\ \midrule
w/o fusion           & 69.8          & 64.5          & \textbf{80.3}          & 89.2          & 79.5          & 74.5          & 63.3          & 65.2          & 76.0          & 29.8          & \textbf{65.3}          & 67.5          & 71.9          & 59.6          & 70.0          & 92.2          & \textbf{63.9}          & 85.2          & 58.6          \\
w/ fusion            & \textbf{70.5} & \textbf{64.8} & \textbf{80.3} & \textbf{89.5} & \textbf{80.5} & \textbf{74.8} & \textbf{63.9} & \textbf{66.0} & \textbf{76.9} & \textbf{30.7} & 64.5 & \textbf{68.3} & \textbf{73.4} & \textbf{61.8} & \textbf{72.4} & \textbf{93.1} & 63.5 & \textbf{85.2} & \textbf{59.3} \\ \midrule
\textbf{Improvement} & \textbf{0.7}  & \textbf{0.3}  & \textbf{0.0}  & \textbf{0.3}  & \textbf{1.0}  & \textbf{0.3}  & \textbf{0.6}  & \textbf{0.8}  & \textbf{0.9}  & \textbf{0.9}  & -0.8 & \textbf{0.8}  & \textbf{1.5}  & \textbf{2.2}  & \textbf{2.4}  & \textbf{0.9}  & -0.4 & \textbf{0.0}  & \textbf{0.7}  \\ \bottomrule
\end{tabular}
\label{tab:back_fusion}
\end{table*}

\vspace{0.5em}\noindent \textbf{Octree $k$-NN.} Based on the two last rows of Table \ref{tab:component_wise}, octree seems only slightly speeds up k-NN from 0.50s to 0.44s. The reason for the slight improvement is that the search space is reduced more significantly on S3DIS by CAPS and late devoxelization (implied by the voxel-to-point ratio in Table \ref{tab:voxel_vs_point}), and thus the runtime difference of $\mathcal{O}(n\log n)$ and $\mathcal{O}(n^2)$ algorithms is not very significant. However, the amount of search space reduction is non-deterministic, highly depending on the point density of different datasets. Octree k-NN is beneficial in reducing the time complexity of the framework: it solves the root cause of the problem by making sure that each component in the framework has a time complexity of less than $\mathcal{O}(n^2)$. For example, in a more challenging case, we can see the effectiveness of octree k-NN more clearly. We report the runtime of the whole scene (without cropping) of STPLS3D dataset with $\sim$4M points in Table \ref{tab:octree}. The results indicate that the runtime of $k$-NN is more significantly reduced from 1.48s to 0.93s.

\vspace{0.5em}\noindent \textbf{Effectiveness of CAPS.} We compare the proposed CAPS with the naive scaling strategy in Table \ref{tab:pyramid_scaling}. While running slightly faster, CAPS achieves AP/AP$_{50}$/AP$_{25}$ of 50.9/67.8/76.9(\%) which is 3.9/3.1/3.2(\%) improvement compared to naive scaling. The qualitative comparison between the two strategies is shown in Figure \ref{fig:pyramid_scaling}. As expected, naive scaling misclassifies the points at the object edges as the background (illustrated in black points), while the proposed CAPS produces more accurate instances.

\begin{table}[]
\centering
\small
\caption{Ablation study with varying point density on S3DIS Area 5. }
\begin{tabular}{l|ccc}
\toprule
Density & AP   & AP$_{50}$ & AP$_{25}$ \\ \midrule
1x      & \textbf{50.9} & \textbf{67.8 }  & \textbf{76.9}   \\
0.5x    & 50.6 & 67.7   & 76.4   \\
0.25x   & 49.2 & 64.5   & 73.8   \\
0.125x  & 46.0   & 62.1   & 72.5   \\ \bottomrule
\end{tabular}
\label{tab:point_density}
\end{table}

\begin{table}[]
\centering
\small
\caption{SoftGroup complements SSTNet for better overall performance on ScanNet validation set.}
\begin{tabular}{l|ccc}
\toprule
Method               & AP            & AP$_{50}$        & AP$_{25}$        \\ \midrule
SoftGroup  & 46.0  & 67.6 & \textbf{78.9} \\
SSTNet \cite{liang2021instance}              & 49.4          & 64.9          & 74.4          \\
SSTNet \cite{liang2021instance} + SoftGroup   & \textbf{51.4} & \textbf{67.7} & 78.2 \\ \bottomrule
\end{tabular}
\label{tab:complement}
\end{table}

\vspace{0.5em}\noindent \textbf{Ablation study on varying point density.} The generalization of the proposed method is demonstrated on various indoor and outdoor datasets with different point densities. We further study the performance of our method with the same dataset and varying point density. To this end, we subsample the original scenes of S3DIS with factors of 0.5x, 0.25x, and 0.125x. Table \ref{tab:point_density} reports the experiment results of SoftGroup++ with varying point density. Compared to the results from the original scenes (indicated by 1x), the performance on 0.5x scenes is approximately on par. As anticipated, as the point density decreases, there is a corresponding decline in performance. Grouping adjacent points becomes more challenging with a reduced point density. Figure \ref{fig:point_density} provides some visual examples to illustrate this point. Despite the low point density, our method generates reasonably accurate instance masks.

	
	

\begin{table}[]
	\centering
	\small
    \caption{Instance segmentation and object detection results on ScanNet v2 validation set. Our method achieves better results on both mask and box AP.}
	\begin{tabular}{l|cc} \toprule
		Method	&	Box	AP$_{50}$	&	Box	AP$_{25}$	\\	\midrule
		F-PointNet	\cite{qi2018frustum}	&	10.8	&	19.8	\\
		GSPN	\cite{yi2019gspn}	&	17.7	&	30.6	\\
		3D-SIS	\cite{hou20193d}	&	22.5	&	40.2	\\
		VoteNet	\cite{qi2019deep}	&	33.5	&	58.6	\\
		3D-MPA	\cite{engelmann20203d}	&	49.2	&	64.2	\\
		PointGroup	\cite{jiang2020pointgroup}	&	48.9	&	61.5	\\
		SSTNet	\cite{liang2021instance}	&	52.7	&	62.5	\\
		HAIS	\cite{chen2021hierarchical}	&	53.1	&	64.3	\\
		\textbf{SoftGroup (ours)}	&	59.4	&	71.6	\\
        \textbf{SoftGroup++ (ours)}	&	\textbf{59.6}	&	\textbf{71.7} \\ \bottomrule
	\end{tabular}
	\label{tab:detection}
\end{table}

\begin{figure*}
	\centering
	\begin{subfigure}{0.22\textwidth}
		\includegraphics[width=\textwidth]{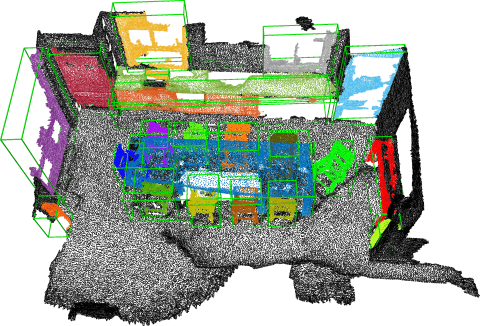}
	\end{subfigure}
	\begin{subfigure}{0.22\textwidth}
		\includegraphics[width=\textwidth]{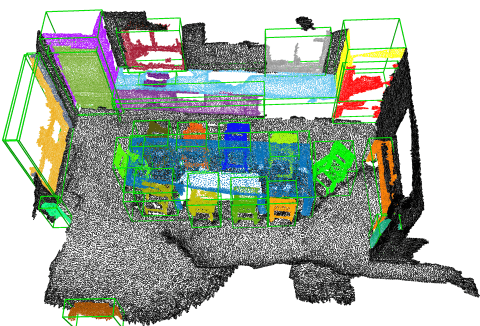}
	\end{subfigure}
	\begin{subfigure}{0.22\textwidth}
		\includegraphics[width=\textwidth]{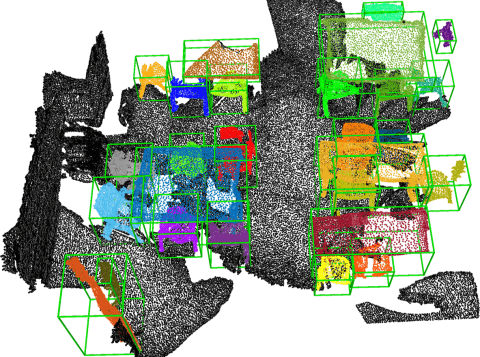}
	\end{subfigure}
	\begin{subfigure}{0.22\textwidth}
		\includegraphics[width=\textwidth]{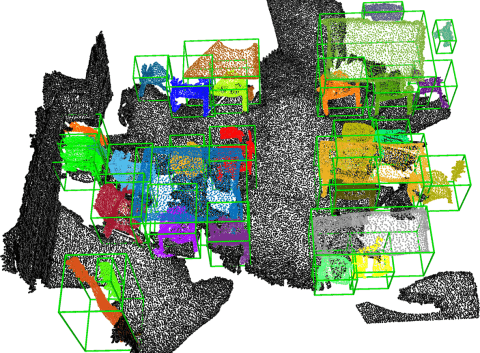}
	\end{subfigure}	
	


	\begin{subfigure}{0.22\textwidth}
		\includegraphics[width=\textwidth]{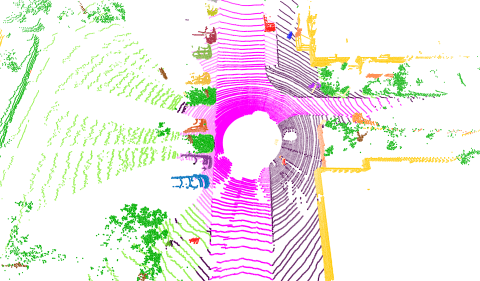}
		\caption*{Prediction}
	\end{subfigure}
	\begin{subfigure}{0.22\textwidth}
		\includegraphics[width=\textwidth]{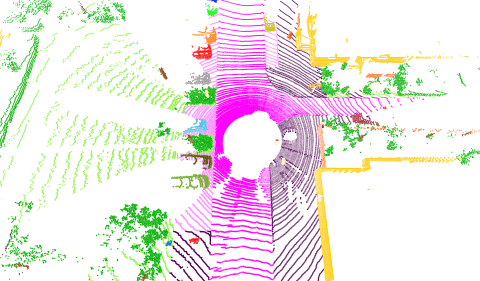}
		\caption*{Ground truth}
	\end{subfigure}
	\begin{subfigure}{0.22\textwidth}
		\includegraphics[width=\textwidth]{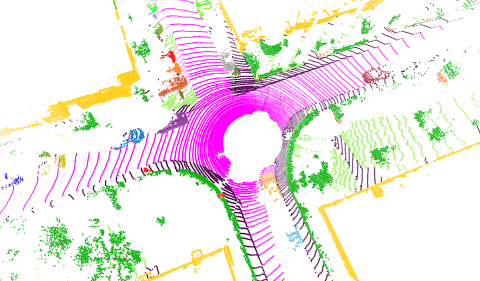}
		\caption*{Prediction}
	\end{subfigure}
	\begin{subfigure}{0.22\textwidth}
		\includegraphics[width=\textwidth]{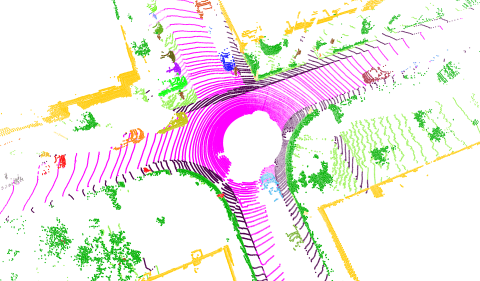}
		\caption*{Ground truth}
	\end{subfigure}
	\caption{\textbf{Visualization examples of object detection on ScanNet v2 (two first rows) and panoptic segmentation on SemanticKITTI (two latter rows).}}
	\label{fig:panoptic}
\end{figure*}

\begin{table*}[]
	\small
        \caption{LiDAR-based panoptic segmentation results on SemanticKITTI test set.}
	\centering
	\begin{tabular}{l|cccc|ccc|ccc|c} \toprule
		Method             & PQ            & PQ$^\dagger$         & RQ            & SQ            & PQ$^{th}$          & RQ$^{th}$          & SQ$^{th}$          & PQ$^{st}$          & RQ$^{st}$          & SQ$^{st}$          & mIoU          \\ \midrule
		RangeNet++ \cite{milioto2019rangenet++} + PointPillars \cite{lang2019pointpillars} & 37.1          & 45.9          & 47.0          & 75.9          & 20.2          & 25.2          & 75.2          & 49.3          & 62.8          & 76.5          & 52.4          \\
		KPConv \cite{thomas2019kpconv} + PV-RCNN \cite{shi2020pv}         & 50.2          & 57.5          & 61.4          & 80.0          & 43.2          & 51.4          & 80.2          & 55.9          & 68.7          & 79.9          & 62.8 \\
		LPASD \cite{milioto2020lidar}                    & 38.0          & 47.0          & 48.2          & 76.5          & 25.6          & 31.8          & 76.8          & 47.1          & 60.1          & 76.2          & 50.9          \\
		PanosterK \cite{gasperini2021panoster}                & 52.7          & 59.9          & 64.1          & 80.7          & 49.4          & 58.5          & 83.3          & 55.1          & 68.2          & 78.8          & 59.9          \\
		4D-PLS \cite{aygun20214d}                    & 50.3          & 57.8          & 61.0          & 81.6          & -             & -             & -             & -             & -             & -             & 61.3          \\
		Panoptic-PolarNet \cite{zhou2021panoptic}        & 54.1          & 60.7          & 65.0          & 81.4          & 53.3          & 60.6          & 87.2 & 54.8          & 68.1          & 77.2          & 59.5          \\
		DS-Net \cite{hong2021lidar}                   & 55.9          & 62.5          & 66.7          & 82.3          & 55.1          & 62.8          & 87.2 & 56.5          & 69.5          & 78.7          & 61.6          \\
        Panoptic-PHNet \cite{li2022panoptic} & \textbf{61.5} & \textbf{67.9} & \textbf{72.1} & \textbf{84.8} & \textbf{63.8} & \textbf{70.4} & \textbf{90.7} & \textbf{59.9} & \textbf{73.3} & \textbf{80.5} & \textbf{66.0} \\ \midrule
		\textbf{SoftGroup (ours)}          & 57.0 & 63.9 & 68.1 & 82.5 & 56.5 & 65.2 & 85.4 & 57.3 & 70.2 & 80.4 & 61.7      \\ 
        \textbf{SoftGroup++ (ours)} & 57.2 & 64.2 & 68.2 & 82.7 & 57.1 & 65.6 & 85.8 & 57.3 & 70.2 & 80.4 & 63.0 \\ \bottomrule   
	\end{tabular}
	\label{tab:panoptic_test}
\end{table*}


\subsection{Generalization}

\vspace{0.5em}\noindent \textbf{Complement to Superpoint-based Instance Segmentation.} Some existing methods, such as STTNet \cite{liang2021instance}, exploit connectivity information among neighboring points to construct superpoint for more accurate instance segmentation. The concept behind SoftGroup is to prevent the error propagation from semantic segmentation to instance segmentation, and we demonstrate that this notion synergistically complements the superpoint-based approach. Table \ref{tab:complement} reports the experimental results of applying SoftGroup on SSTNet. SoftGroup brings substantial performance gain on SSTNet in terms of AP/AP$_{50}$/AP$_{25}$ from 49.4/64.9/74.4(\%) to 51.4/67.7/78.2(\%). Compared to original SoftGroup, the combination of SoftGroup and SSTNet achieves significant AP improvement of 5.4\%.

\vspace{0.5em}\noindent \textbf{Semantic Segmentation with Back Fusion.} We improve the semantic segmentation performance by fusing back the instance masks to semantic maps. To handle overlapping, the instances masks are pasted onto the semantic maps in ascending order of their confident scores. Table \ref{tab:back_fusion} reports the ablation results of back fusion. Overall, back fusion achieves mIoU improvement of 0.7 points on ScanNet v2 validation set. Notably, back fusion improves the performance of classes with close semantic meaning, such as \texttt{cabinet} v.s. \texttt{fridge}, \texttt{curtain} v.s. \texttt{shower curtain}.

\vspace{0.5em}\noindent \textbf{Generalization to Object Detection.} To obtain object detection results, we follow the approach in \cite{engelmann20203d} to extract a tight axis-aligned bounding box from the predicted instance masks. Table \ref{tab:detection} report the object detection results on ScanNet v2 validation. The proposed SoftGroup and SoftGroup++ significantly outperform the other methods in terms of bounding box AP. SoftGroup++ achieves the best overall performance with AP$_{50}$/AP$_{25}$ of 59.6/71.7(\%).

\vspace{0.5em}\noindent \textbf{Generalization to Panoptic Segmentation.} Since the proposed method produces both semantic and instance segmentation results, it naturally generalizes to panoptic segmentation. Table \ref{tab:panoptic_test} reports the panoptic segmentation results on SemanticKITTI dataset. The proposed SoftGroup and SoftGroup++ achieve competitive performance. Panoptic-PHNet attains the best overall results with PQ of 61.5. Panoptic-PHNet employs fused 2D-3D backbone, transformer-based offset branch, and center grouping module, which leads to significant performance gain. We posit that the SoftGroup concept is not only orthogonal to the advancements seen in Panoptic-PHNet, but it also complements them. This suggests potential for additional performance improvements. However, the lack of publicly released implementation details for Panoptic-PHNet makes the reproduction of the reported results a significant challenge. Thus, we defer a more thorough exploration of this potential performance enhancement to future research.

Figure \ref{fig:panoptic} illustrates the visualization of object detection and panoptic segmentation outputs. Our method produces plausible predictions under different scenarios. In summary, quantitative and qualitative results on different tasks and datasets demonstrate the versatility and generality of the proposed method.

\section{Conclusion}
We present SoftGroup and its extended SoftGroup++ for accurate and scalable instance segmentation on 3D point clouds. SoftGroup performs grouping on soft semantic scores to address the problem stemming from hard grouping on locally ambiguous objects. Then a top-down refinement stage is constructed to refine the positive samples and suppress the negatives. To efficiently process large-scale scenes, SoftGroup++ is introduced for low time complexity and search space. Octree $k$-NN replaces vanilla $k$-NN to reduce the time complexity from $\mathcal{O}(n^2)$ to $\mathcal{O}(n\log n)$. Class-aware pyramid scaling and late devoxelization reduce the search space and runtime of intermediate components. Extensive experiments on various datasets demonstrate the superiority and generality of our method. 

\ifCLASSOPTIONcompsoc
  \section*{Acknowledgments}
\else
  \section*{Acknowledgment}
\fi
This work was partly supported by Institute for Information communications Technology Planning Evaluation (IITP) grant funded by the Korea government (MSIT) (2021-0-01381, Development of Causal AI through Video Understanding, and partly supported by Institute of Information \& Communications Technology Planning \& Evaluation (IITP) grant funded by the Korea government (MSIT) (No. 2019-0-01371, Development of brain-inspired AI with human-like intelligence).

\bibliographystyle{IEEEtran}
\bibliography{refs}

\vskip -0.5in
\begin{IEEEbiography}[{\includegraphics[width=1in,height=1.25in,clip,keepaspectratio]{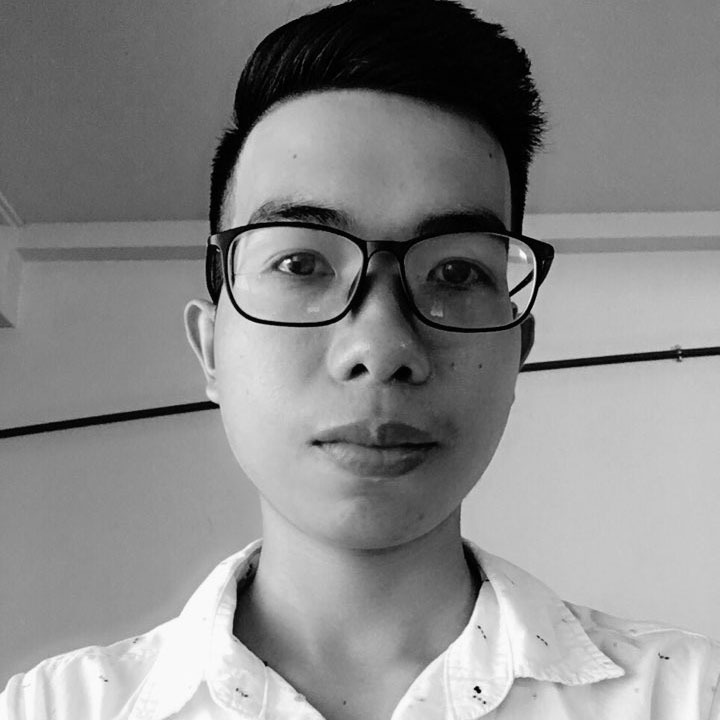}}]{\textbf{Thang Vu}} received B.Sc degree in electronic and telecommunication engineering from Hanoi University of Science and Technology, in 2016, and the M.Sc. degree in electrical engineering from Korea Advanced Institute of Science and Technology, in 2019. He is currently pursuing a Ph.D. degree with Korea Advanced Institute of Science and Technology. His research interests include object detection and instance segmentation on 2D/3D data, sensor fusion, and various recognition problems.
\end{IEEEbiography}

\vskip -0.45in
\begin{IEEEbiography}[{\includegraphics[width=1in,height=1.25in,clip,keepaspectratio]{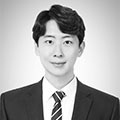}}]{\textbf{Kookhoi Kim}} received B.Sc degree in electronic engineering from Hanyang University, in 2020, and the M.Sc. degree in electrical engineering from Korea Advanced Institute of Science and Technology, in 2022. His research interests include machine learning and deep learning for computer vision.
\end{IEEEbiography}

\vskip -0.6in
\begin{IEEEbiography}[{\includegraphics[width=1in,height=1.25in,clip,keepaspectratio]{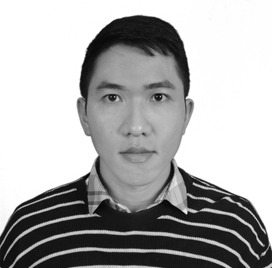}}]{\textbf{Tung M. Luu}} received B.Sc degree in electronic and telecommunication engineering from Hanoi University of Science and Technology, in 2017, and the M.Sc. degree in electrical engineering from Korea Advanced Institute of Science and Technology, in 2020. He is currently pursuing a Ph.D. degree with Korea Advanced Institute of Science and Technology. His research interests include machine learning, deep learning, and reinforcement learning.
\end{IEEEbiography}

\vskip -0.6in
\begin{IEEEbiography}[{\includegraphics[width=1in,height=1.25in,clip,keepaspectratio]{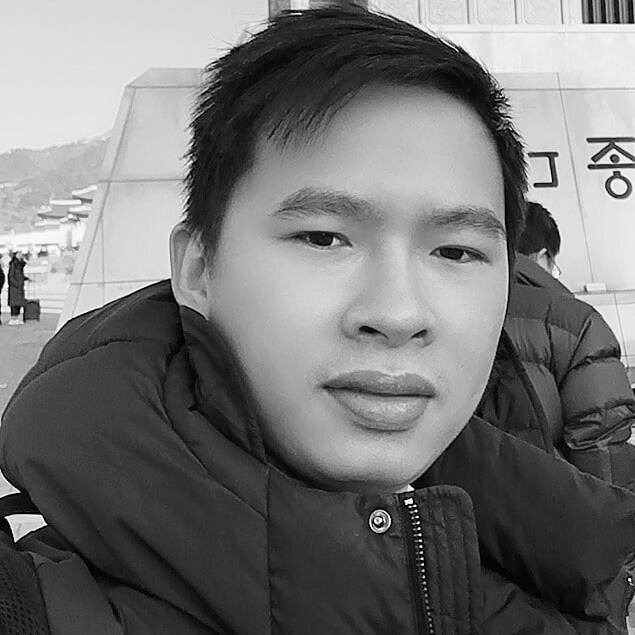}}]{\textbf{Thanh Nguyen}} received B.Sc degree in electronic and automation engineering from Ho Chi Minh City University of Science and Technology, in 2015. He has been pursuing an M.Sc. and Ph.D. degree in Korea Advanced Institute of Science and Technology since 2018. His research interests include machine learning, deep learning, and reinforcement learning.
\end{IEEEbiography}

\vskip -0.6in
\begin{IEEEbiography}[{\includegraphics[width=1in,height=1.25in,clip,keepaspectratio]{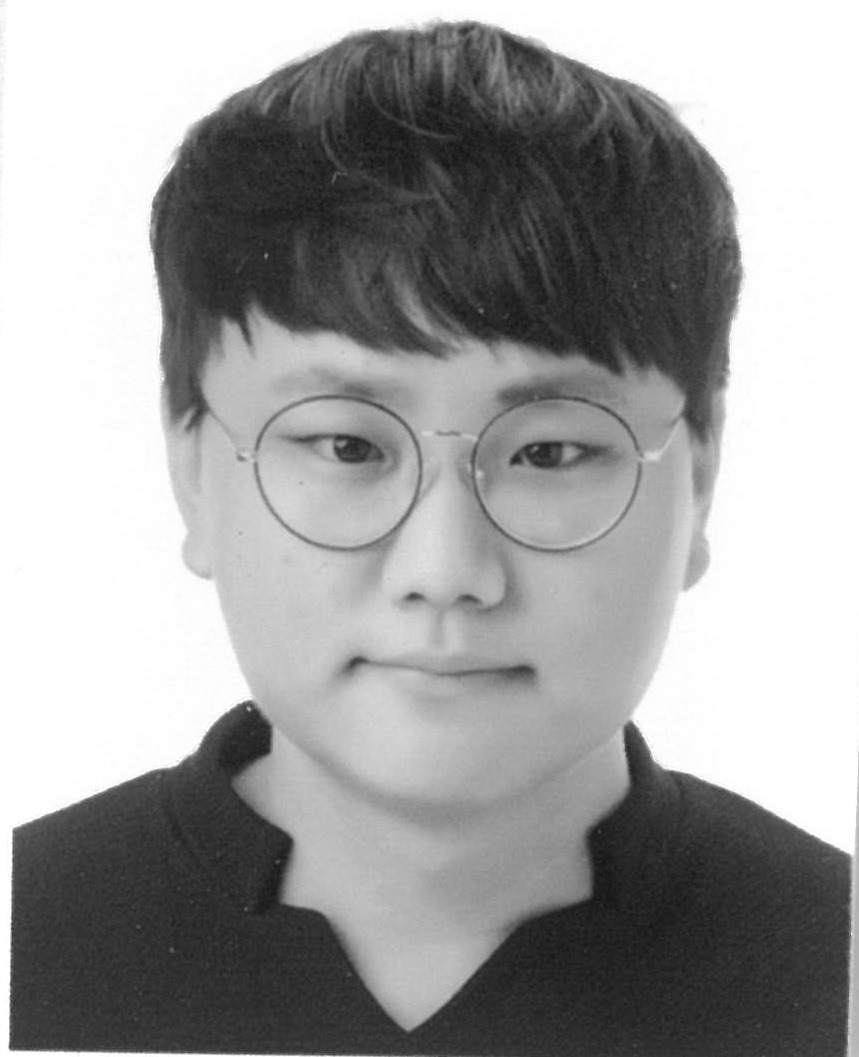}}]{\textbf{Junyeong Kim}} received the B.S., M.S., and Ph.D. degrees from KAIST, Republic of Korea, in 2021. He was a post-doctoral researcher associate in the Artificial Intelligence and Machine Learning Laboratory, School of Electrical Engineering, KAIST, Republic of Korea. He currently works as an assistant professor, Department of AI, Chung-Ang University, Republic of Korea. His research interests include visual-language reasoning, visual question answering, and various video-based problem.
\end{IEEEbiography}

\vskip -0.45in	
\begin{IEEEbiography}[{\includegraphics[width=1in,height=1.25in,clip,keepaspectratio]{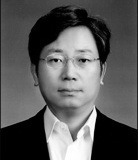}}]{\textbf{Chang D. Yoo}} (Senior Member, IEEE) received the BS degree in engineering and applied science from California Institute of Technology, the MS degree in electrical engineering from Cornell University, and the PhD degree in electrical engineering from the Massachusetts Institute of Technology. From 1997 to 1999, he was a senior researcher with Korea Telecom (KT). Since 1999, he has been on the faculty with the Korea Advanced Institute of Science and Technology (KAIST), where he is currently a professor in the School of Electrical Engineering and an adjunct professor with the School of Computer Science. He served as Associate Editor for IEEE/ACM Transactions on Audio Speech and Language Processing, IEEE Transactions on Information Forensics and Security and IEEE Signal Processing Letters. He also served as Dean of the Office of Special Projects and Dean of the Office of International Relations, respectively.
\end{IEEEbiography}




\end{document}